\documentclass[lettersize,journal]{IEEEtran}
\usepackage{amsmath,amsfonts}
\usepackage{algorithmic}
\usepackage{algorithm}
\usepackage{array}
\usepackage[caption=false,font=normalsize,labelfont=sf,textfont=sf]{subfig}
\usepackage{textcomp}
\usepackage{stfloats}
\usepackage{url}
\usepackage{verbatim}
\usepackage{graphicx}
\usepackage{cite}
\usepackage[numbers,sort&compress]{natbib}
\usepackage{enumerate}
\usepackage{multirow}
\usepackage{color, booktabs}
\usepackage{balance}
\usepackage{colortbl, xcolor}
\begin{document}

\title{Quality Assessment and Distortion-aware Saliency Prediction for AI-Generated Omnidirectional Images}
\author{\IEEEauthorblockN {Liu Yang, Huiyu Duan, Jiarui Wang, Jing Liu, Menghan Hu, Xiongkuo Min, \textit{Member}, \textit{IEEE}, \\
Guangtao Zhai, \textit{Fellow}, \textit{IEEE} and Patrick Le Callet, \textit{Fellow}, \textit{IEEE}
}
\thanks{Liu Yang, Huiyu Duan, Jiarui Wang, Xiongkuo Min, Guangtao Zhai are with the Institute of Image Communication and Network Engineering, Shanghai Jiao Tong University, Shanghai 200240, China (e-mail: ylyl.yl@sjtu.edu.cn; huiyuduan@sjtu.edu.cn; wangjiarui@sjtu.edu.cn; minxiongkuo@sjtu.edu.cn; zhaiguangtao@sjtu.edu.cn).

Jing Liu is from Tianjin University, Tianjin 300072, China (e-mail: jliu\_tju@tju.edu.cn).

Menghan Hu is from East China Normal University, Shanghai 200241, China (e-mail: mhhu@ce.ecnu.edu.cn).

Patrick Le Callet is with the Polytech Nantes, Université de Nantes, 44306 Nantes, France (e-mail: patrick.lecallet@univ-nantes.fr).
}}

% \vspace{-20pt}
\maketitle
\begin{abstract}
With the rapid advancement of Artificial Intelligence Generated Content (AIGC) techniques, AI generated images (AIGIs) have attracted widespread attention, among which AI generated omnidirectional images (AIGODIs) hold significant potential for Virtual Reality (VR) and Augmented Reality (AR) applications. AI generated omnidirectional images exhibit unique quality issues, however, research on the quality assessment and optimization of AI-generated omnidirectional images is still lacking. To this end, this work first studies the quality assessment and distortion-aware saliency prediction problems for AIGODIs, and further presents a corresponding optimization process. Specifically, we first establish a comprehensive database to reflect \underline{\textit{h}}uman \underline{\textit{f}}eedback for AI-generated \underline{\textit{o}}mnidirectionals, termed OHF2024, which includes both subjective quality ratings evaluated from three perspectives and distortion-aware salient regions. Based on the constructed OHF2024 database, we propose two models with shared encoders based on the BLIP-2 model to evaluate the human visual experience and predict distortion-aware saliency for AI-generated omnidirectional images, which are named as BLIP2OIQA and BLIP2OISal, respectively. Finally, based on the proposed models, we present an automatic optimization process that utilizes the predicted visual experience scores and distortion regions to further enhance the visual quality of an AI-generated omnidirectional image.
Extensive experiments show that our BLIP2OIQA model and BLIP2OISal model achieve state-of-the-art (SOTA) results in the human visual experience evaluation task and the distortion-aware saliency prediction task for AI generated omnidirectional images, and can be effectively used in the optimization process. The database and codes will be released on {https://github.com/IntMeGroup/AIGCOIQA} to facilitate future research.
\end{abstract}

\vspace{-12pt}
\section{Introduction}
Omnidirectional images (ODIs) can provide 360-degree free viewing experience, which are widely used in the fields of Virtual Reality (VR), Augmented Reality (AR), game development, cultural heritage protection, \textit{etc.}
ODIs are generally shown in the Equirectangular Projection (ERP) format , which warps an image to adapt to the omnidirectional characteristic. Artificial Intelligence Generated Content (AIGC) refers to generating various forms of content such as texts, images, musics, videos, and 3D interactive contents, \textit{etc.}, using AI. Thanks to the rapid advancement of generative models such as Generative Adversarial Network (GAN) \cite{gan}, Variational Auto Encoders (VAE) \cite{vae} and Diffusion Models (DMs) \cite{dm}, \textit{etc.}, as well as language-vision models such as CLIP \cite{clip} and BLIP-2 \cite{blip2}, \textit{etc.}, AI Generated Images (AIGIs) have attracted widespread attention. Recently, many omnidirectional image generation models \cite{mvdiffusion, text2light} have also been proposed and have shown great potential for fast and creative omnidirectional content generation.

\begin{figure}[!t]
\centering
\includegraphics[width=3.5in]{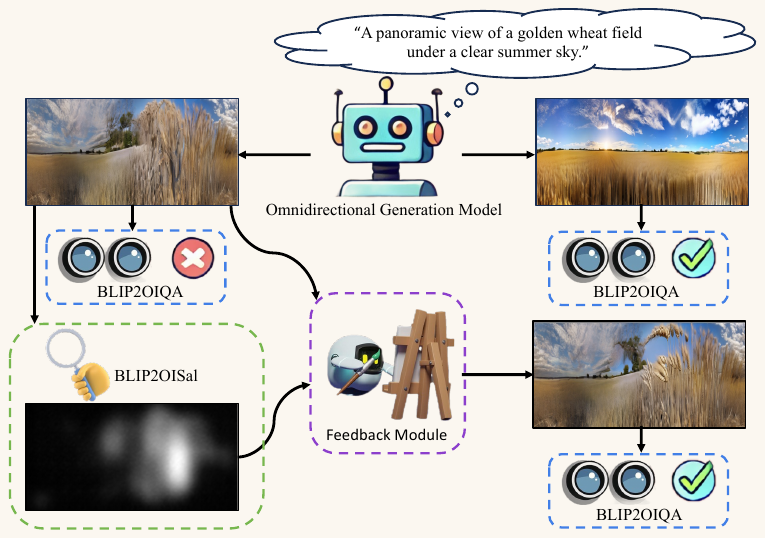}
\vspace{-18pt}
% \caption{Our BLIP2OIQA and BLIP2OISal models help understanding and improving the quality of AI generated omnidirectional images for real-world applications through the following workflow: High-quality images, as predicted by BLIP2OIQA based on human visual preference, can be directly outputted. For low-quality images, BLIP2OISal performs distortion-aware saliency prediction, guiding local quality modifications through an inpainting model. This process results in significant quality enhancement of the images, making them suitable for application.}
\caption{Workflow of our proposed optimization process.
For the generated omnidirectional images, we first employ the BLIP2OIQA model for human visual experience evaluation. Images of satisfactory quality are directly outputted, while those of unsatisfactory visual quality undergo further optimization process. The BLIP2OISal model predicts distortion-aware salient regions, which are then fed into the feedback module with the unsatisfactory images to guide the optimization process.
The refined images are then re-evaluated by BLIP2OIQA to ensure that the output image is satisfactory.}
\vspace{-18pt}
\label{fig:workflow}
\end{figure}

Although omnidirectional image generation models have many potential applications \cite{dreamscene360, 4k4dgen}, current AI-generated omnidirectional images (AIGODIs) may not meet the visual expectations of the users. Compared with natural omnidirectional images, AI generated omnidirectional images may not only have quality problems such as blur, noise, lighting issues, \textit{etc.}, but also exhibit some unique degradations caused by the generation, such as unrealism, unreasonable composition, and low relevance between the prompt and image. These degradation issues can significantly affect the immersive experience of the users in real-world applications. Therefore, it is significant and necessary to evaluate the visual experience of AI-generated omnidirectional images, identify regions with severe distortion, and either discard or optimize them accordingly.

While there are quantitative metrics such as Fréchet Inception Distance (FID) \cite{fid} and Inception Score (IS) \cite{IS} proposed to evaluate the performance of generation models, these algorithms can only assess the authenticity dimension of an image set, while lacking the capability to evaluate the quality of a single image and perform the text-image correspondence measure. Some other metrics, like CLIPScore \cite{clipscore}, mainly consider the text-image correspondence dimension of AI-generated images, while ignoring the comprehensive assessment of authenticity and overall quality of omnidirectional images. Recently, some IQA studies \cite{aigciqa,li2023agiqa} have proposed databases and models for evaluating human visual preference for common AIGIs. However, AI-generated omnidirectional images are significantly different with AI-generated common images in terms of characteristic and applications, thus previous IQA studies for common AIGIs can not generalize well on AIGODIs. Moreover, these works only study the human preference ranking problem, while lacking research on the distortion region prediction problem, which is of great significance for enhancing the visual experience of AI generated omnidirectional images.

Visual attention analysis and prediction have been important tasks in multimedia and computer vision research for a long time, which can significantly help image quality evaluation algorithms and image enhancement algorithms \cite{saliency-iqa, saliency-enhancement}. 
% Generating images with higher quality in regions where the human eyes pay more attention to significantly enhances the overall visual experience. This is particularly important for omnidirectional images, where the immersive experience necessitates higher quality in the salient regions of the image. Therefore, prediction of distortion regions in saliency regions of images is important for understanding the quality of generated images, and distortion-aware saliency prediction is instructive for modification of AI generated omnidirectional images. However, there is no work to accomplish this task.
For AI-generated images, it is important to predict distortion-aware salient regions rather than typical visual salient regions, which can significantly guide the corresponding image optimization process \cite{liang2024rich}.
Although many previous works have studied the visual saliency prediction problem for natural omnidirectional images \cite{mc360, ref5}, visually salient regions can be of both high quality or low quality.
Therefore, here we define a new task termed distortion-aware saliency predicting, aiming to predict visually salient areas with strong degradations, which can be effectively used in the optimization process.

In this paper, we present the first work of assessing visual experience, predicting distortion-aware saliency, conducting image optimization for AI-generated omnidirectional images. Specifically, we first construct a database which contains rich \underline{\textit{h}}uman \underline{\textit{f}}eedback for AI generated \underline{\textit{o}}mnidirectional images (OHF2024). 
As shown in Fig. \ref{fig:dataset}, Our OHF2024 database contains 600 omnidirectional images generated using five models based on 50 diverse scene prompts, 1,800 Mean Opinion Scores (MOSs) from three perspectives including quality, comfortability and correspondence, and 600 distortion-aware saliency maps derived from human annotations.
% which contains 600 AI generated omnidirectional images and corresponding Mean Opinions Scores (MOSs) from three perspectives named quality, comfortability and correspondence. We also collect 1200 detailed annotations of distortion regions, resulting in a total of 600 distortion-aware saliency maps. 
Furthermore, we propose two models with shared-weight encoders to evaluate human visual experience and predict distortion-aware saliency for AIGODIs based on the pre-trained BLIP-2 \cite{blip2} model, which are named as BLIP2OIQA and BLIP2OISal, respectively. Our proposed models achieve state-of-the-art (SOTA) performance in human visual experience evaluation and distortion-aware saliency prediction tasks for AIGODIs.
% BLIP2OIQA, an IQA model which achieves the state-of-the-art (SOTA) performance on evaluating human visual preference from all three perspectives we propose for AI generated omnidirectional images in OHF2024 database. To obtain a richer understanding of the quality of AI generated omnidirectional images, we also propose the BLIP2OISal model, which achieves the state-of-the-art performance on distortion-aware saliency prediction.
% Finally, we demonstrate that rich quality information gained from both the BLIP2OIQA and BLIP2OISal models can be effectively fed back into AI generated omnidirectional images. This feedback not only filters out AI generated omnidirectional images that do not meet human visual preferences but also helps refine the images based on the salient distorted regions predicted by BLIP2OISal, thereby producing omnidirectional images that better align with human visual preferences.
Finally, we present an automated optimization process for AIGODIs as shown in Fig. \ref{fig:workflow} and demonstrate that the visual quality of AI-generated omnidirectional images can be significantly improved based on human feedback predicted by our proposed models.

The main contribution of this paper can be summarized as follows:
\begin{itemize}
\item{We construct the first comprehensive database containing rich human feedback for AI generated omnidirecional images, named OHF2024, including both human visual experience ratings and distortion-aware saliency.}
% \item{We propose the BLIP2OIQA model, which is the first to utilizes attention modules to thoroughly extract inter-viewport text-image-fused features and facilitate cross-viewport feature aggregations. Experiment have shown that our BLIP2OIQA model achieves the state-of-the-art performance in analyzing human visual preferences for AI generated omnidirectional images from our proposed evaluation perspectives include quality, comfortability and correspondence.}
% \item{We extend the task of image saliency to predicting salient distorted regions of AI generated omnidirectional images. We propose the BLIP2OISal model, which employs a feature fusion module to effectively integrate spatial information with text-image-fused features, and achieves the state-of-the-art performance in distortion-aware saliency prediction task.}
% \item{We demonstrate that both the BLIP2OIQA and BLIP2OISal models can be effectively fed back into applications of AI generated omnidirectional images. While BLIP2OIQA model helps to filter out AI generated omnidirectional images that do not align with human visual preferences, BLIP2OISal model plays an instructive role in refining images based on predicted salient distortion-aware regions, thereby enhancing the efficiency of AI generated omnidirectional images in practical applications.}
 \item{We propose two models with a shared-wise encoder based on the BLIP-2 model, termed BLIP2OIQA and BLIP2OISal, which are equipped with a quality decoder and a saliency decoder to perform visual experience assessment and distortion-aware saliency, respectively.}
 \item{We propose an automatic optimization process based on the proposed BLIP2OIQA and BLIP2OISal models.}
 \item{Experimental results demonstrate that our models achieve state-of-the-art performance in human visual experience evaluation and distortion-aware saliency prediction on AIGODIs, respectively, and the automatic optimization process can effectively help enhance the visual quality of an AIGODI.}
 % \item{On the decoder side, BLIP2OIQA is the first to utilize attention modules for thoroughly extracting inter-viewport text-image-fused features and facilitating cross-viewport feature aggregation to generate quality representations, while BLIP2OISal employs a well-designed feature fusion module to effectively integrate spatial information with text-image-fused features. Experimental results demonstrate that our models achieve state-of-the-art performance in human visual experience evaluation and distortion-aware saliency prediction on AIGODIs, respectively.}
\end{itemize}
\vspace{-15pt}
\section{Related Work}
\subsection{Image Quality Assessment}
Numerous databases and models have been proposed for Image Quality Assessment (IQA) to study human visual preferences for images.
\subsubsection{IQA databases and models for natural images}\label{sec:Databases and Models}
Due to the source and degradation difference, various IQA databases and models have been proposed for classical 2D images \cite{koniq, spaq}.
Omnidirectional image quality assessment (OIQA) is a unique branch in the field of IQA, as omnidirectional images differ significantly from classical images in terms of format, viewing mode, and distortion characteristics, and has important applications in VR and AR. Some previous works \cite{ref2, mc360} have constructed OIQA databases and proposed OIQA models to study human visual preference for omnidirectional images.
Some studies have also explored the immersive experience problem. For instance, Duan \textit{et al.} \cite{itu} construct the CFIQA database and develop the CFIQA model to objectively evaluate the quality of AR images. Zhu \textit{et al.} \cite{esiqa} construct the ESIQA database and conduct a well-organized subjective experiment to study human visual preference for images captured by Vision Pro.

\subsubsection{IQA databases and models for AI-generated images}
With the advancement of AIGC technology, there has been a significant growth of IQA datasets and models specifically proposed for AI generated images. Li \textit{et al.} \cite{li2023agiqa} have constructed the first large-scale database of AI generated images and proposed an IQA model StairReward accordingly. Wang \textit{et al.} \cite{aigciqa} have proposed to evaluate human visual preference for AIGIs from three perspectives including quality, authenticity and correspondence, and construct the AIGCIQA2023 database. They have also proposed MINT-IQA model \cite{aigciqa_model} to better understand and explain human visual preference utilizing vision-language instruction tuning strategy. 

However, studies on AIGODIs are still lacking. As stated in Section \ref{sec:Databases and Models}, IQA of natural omnidirectional images is distinct from that of 2D natural images, thus it is also important to study the IQA problem for AI-generated omnidirectional images.
\vspace{-6pt}
\subsection{Saliency Prediction}
Human visual attention analysis and prediction have always been crucial tasks. In order to understand visual attention behaviour, some saliency databases \cite{salicon, mit1003, cat2000} have been constructed. Many models have been proposed for saliency prediction, which can be broadly categorized into classical models and deep learning-based models. 
Classical models, such as AIM \cite{aim}, SMVJ \cite{smvj}, CovSal \cite{covsal} primarily extract simple low-level features such as intensity and color from images and aggregate them to generate saliency maps. Deep learning-based models \cite{ref4, salicon1, sam} are able to capture more comprehensive visual information from images, thus yielding superior performance in saliency prediction tasks.

\begin{figure*}[!t]
\vspace{-18pt}
\centering
\includegraphics[width=6.5in]{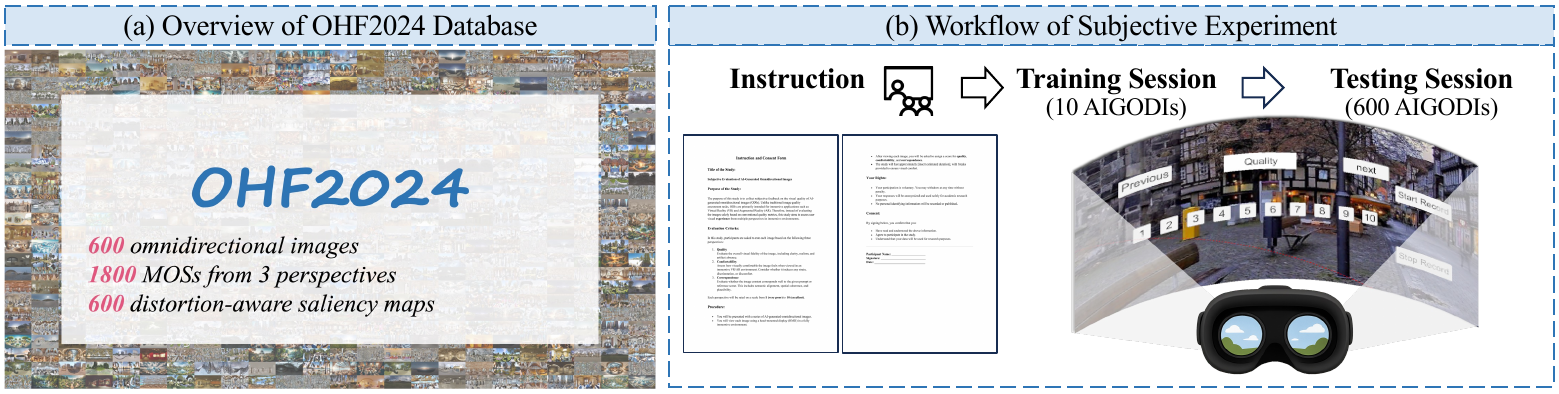}
\vspace{-12pt}
\caption{Overview and subjective experiment procedure of our OHF2024 dataset.}
\vspace{-18pt}
\label{fig:dataset}
\end{figure*}

Saliency prediction task can be applied in various multimedia \cite{autism2, autism3, survey} and computer vision tasks \cite{BIQA, autism4}. However, the research that integrates saliency prediction with distortion prediction to provide a deeper understanding for the local quality of an image is still lacking. This motivates us to develop a saliency model specifically for predicting distortion-aware salient regions in a generated image.
\vspace{-6pt}
\section{OHF2024 Database}
As shown in Fig. \ref{fig:dataset}, we first establish a large-scale database containing rich human feedback for AI-generated omnidirectional images. Unlike previous single-purpose IQA or saliency databases,
our database contains both subjective ratings and distortion-aware saliency to aid in understanding the quality of AI generated omnidirectional images. 
% our database contains both subjective ratings and distortion-aware saliency to help better understand human visual preferences.
Detailed procedures for image collection and subjective experiment are described as follows.
\vspace{-9pt}
\subsection{Image Collection}\label{sec:collection}
Due to the absence of omnidirectional image generation prompts, We first collect 50 omnidirectional images from the MVDiffusion \cite{mvdiffusion,matterport} training dataset and SUN360 \cite{sun360} as natural instances. Then we use BLIP-2 \cite{blip2} to annotate the images to form preliminary prompts, and use GPT4 to add details and polish them. The final prompts describe 25 indoor and 25 outdoor scenes, respectively, which are diverse and cover a wide range of scenarios. These prompts include rich scene details, enhancing the text-image correspondence and making the generated images more distinguishable.

For each prompt, we adopt 5 omnidirectional image generation models, including MVDiffusion \cite{mvdiffusion}, Text2Light \cite{text2light}, DALLE \cite{dalle}, omni-inpainting \cite{mvdiffusion, inpainting} and a fine-tuned Stable Diffusion model \cite{sd} to generate omnidirectional images. When fine-tuning Stable Diffusion model \cite{sd}, we use a total of 6000 BLIP-2-labeled \cite{blip2} indoor/outdoor equirectangular-projected (ERP) omnidirectional images to only fine-tune the Unet module to achieve indoor/outdoor omnidirectional generation, respectively. We generate two omnidirectional images for each of the first four generation models  mentioned above, respectively, and one for fine-tuned Stable Diffusion model \cite{sd}. For MVDiffusion \cite{mvdiffusion} generation, we also adjust the denoising time to generate two distinguishable images for each prompt. Overall, taking natural omnidirectional images into account, we have a total of 12 omnidirectional images for each prompt, and a total of $(25+25)\times 12 = 600$ omnidirectional images in the database.
Fig. \ref{fig:dataset} (a) gives an overview of the images in our OHF2024.

\vspace{-12pt}
\subsection{Subjective Experiment Setup}\label{sec:subjective-experiment}
Based on the collected AIGODIs, we further conduct a subjective experiment to obtain a rich understanding of the quality of AI-generated omnidirectional image from both human visual experience scores and annotations of salient distorted regions. 

First of all, subjects are asked to provide subjective quality scores for AI generated omnidirectional images. Due to the inherent characteristic of omnidirectional images generated by AI, human visual experience for AI-generated images is generally evaluated from multiple dimensions rather than only from the dimension of ``quality'', as discussed in AIGCIQA2023 \cite{aigciqa}. However, different from the general AIGC IQA problem \cite{aigciqa}, AI generated omnidirectional images are mainly produced for VR and AR applications, where users \textit{experience}, rather than simply \textit{view} the images in a fully immersive environment.  Consequently, ODIs exhibit particular visual characteristics \cite{ref1,ref2,ref3,ref4,ref5,ref6,ref7,ref8}. Therefore, in this paper, we propose to evaluate human visual preferences for generated omnidirectional images from three perspectives, including quality, comfortability and correspondence.

% \vspace{-3pt}
The first evaluative dimension is ``\textbf{quality}'', \textit{i.e.}, a comprehensive score of low-level visual qualities such as color, lighting and clarity, \textit{etc}. Since generated omnidirectional images are presented with immersive experience, we further present the second evaluation perspective, \textit{i.e.}, ``\textbf{comfortability}'', which is defined as the users' experiencing preference for AI-generated omnidirectional images. Specifically, subjects are asked to give an overall score for the authenticity level, structure deformation level, as well as perceived comfort. Low comfortability ratings indicate an uncomfortable viewing experience that may even induce sickness.
Since the generated omnidirectional images are mainly produced via the control of the text prompts, the correspondence between text and image is also a critical criteria for evaluating the quality of AI-generated omnidirectional images, \textit{i.e.}, ``text-image \textbf{correspondence}''.

\vspace{0pt}
Then, a total of 20 subjects (10 male and 10 female, aged 18 to 30) are asked to view 600 omnidirectional images and score them from the perspectives of quality, comfortability and correspondence ranging from 1 to 10 with an interval of 1 based on the subjective perception. All subjects have normal or corrected-to-normal vision. Each perspective for each image is rated by at least 20 subjects. We use a HTC VIVE Pro Eye as the head-mounted display (HMD). The images are randomly sorted and displayed sequentially based on the software designed using the Unity, as shown in Fig. \ref{fig:dataset} (b). Subjects are first instructed on the experiment and evaluation dimensions, followed by a training session with 10 AIGODIs to ensure correct understanding and familiarity with the scoring process. The official testing session then begins. To mitigate discomfort or sickness associated with prolonged HMD use, subjects are provided with a 10-minute break after every hour of viewing.

\vspace{0pt}
Moreover, to gain deeper insights into the quality issues of AI-generated omnidirectional images from a distortion-aware saliency perspective, we instruct each subject to provide voice descriptions and handle-click direction for regions with severe distortions.
The handle-click directions point out the distorted regions within the images, while the descriptions describe the specific distortions, including low-level distortions such as blur, noise, \textit{etc.}, as well as high-level distortions such as deformation and warping, \textit{etc.}

\begin{figure*}[!t]
\vspace{-21pt}
\centering
\includegraphics[width=6.5in]{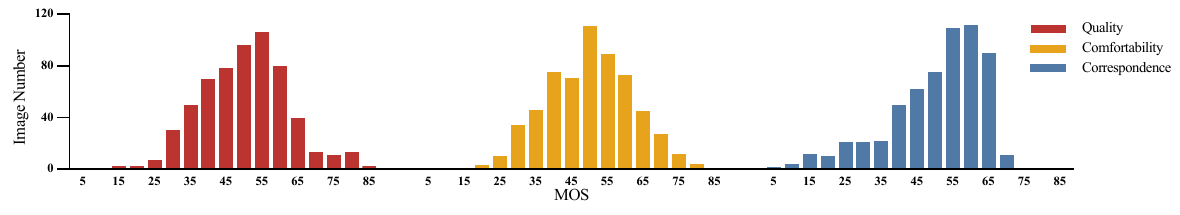}
\vspace{-12pt}
% \caption{(a) MOS distribution of quality score. (b) MOS distribution of comfortability score. (c) MOS distribution of correspondence score.}
\caption{MOS distribution of quality, comfortability and correspondence dimensions.}
\vspace{-6pt}
\label{fig:mos}
\end{figure*}

\begin{figure*}[!t]
\centering
\includegraphics[width=6.5in]{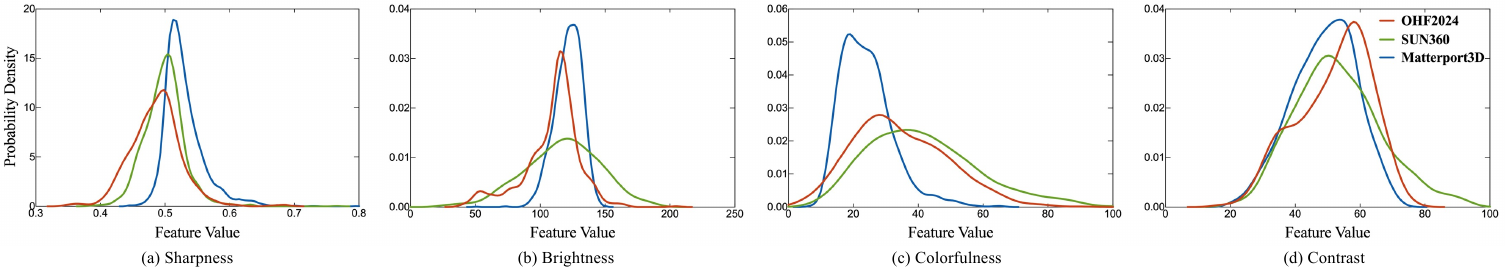}
\vspace{-12pt}
\caption{Kernel distribution of four selected features of three databases: OHF2024, SUN360 \cite{sun360}, Matterport3D \cite{matterport}.}
\vspace{-15pt}
\label{fig:kernel}
\end{figure*}

\begin{figure}[!t]
\vspace{-3pt}
\centering
\includegraphics[width=3.0in]{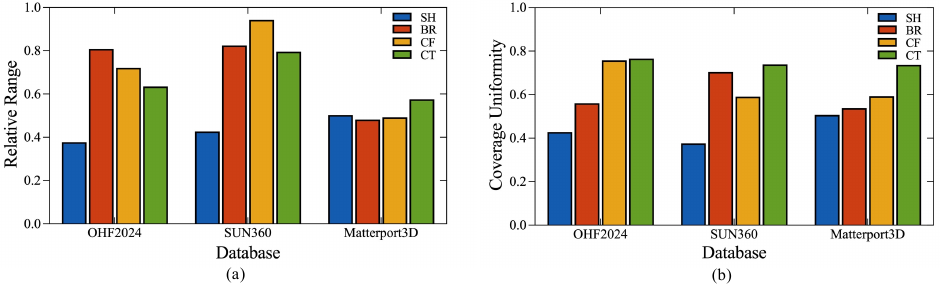}
\vspace{-12pt}
\caption{(a) Relative range and (b) Coverage uniformity of the four selected features on the three databases.}
\vspace{-15pt}
\label{fig:UR}
% \vspace{-12pt}
\end{figure}
\vspace{-12pt}
\subsection{Subjective Data Processing}
\subsubsection{IQA data processing}
We follow the instructions of ITU \cite{ITU-R_BT_500-13} to conduct the outlier detection and subject rejection. As a result, no subjects are rejected and the rejection ratio is 3$\%$ for all ratings. For the remaining valid subjective scores, we convert the raw ratings into Z-scores, then linearly scale them to the [0,100] range. The final 
Mean Opinion Score (MOS) is calculated as follows: 
\vspace{-3pt}
\begin{gather}
z_{ij}=\frac{m_{ij}-\mu_{i}}{\sigma_{i}},\quad z'_{ij}=\frac{100z_{ij}+3}{6}\\
\vspace{-6pt}
MOS_{j}=\frac{1}{N}\sum_{i=1}^{N}z'_{ij}
\vspace{-6pt}
\end{gather}
% \vspace{-3pt}
where $m_{ij}$ is the subjective score given by the $i$-th subject to the $j$-th image, $\mu$ is the mean score given by the $i$-th subject, $\sigma$ is the standard deviation and $N$ is the total number of subjects.

Fig. \ref{fig:mos} illustrates histograms of the MOSs of quality, comfortability, and correspondence, respectively. The MOSs distribution shows that our database encompasses a broad range of human visual experience scores, indicating its diversity. Moreover, different perspectives have different distributions, which also illustrates the differences between the three evaluation perspectives.

\subsubsection{Distortion-aware saliency data processing}
Human visual attention is significantly influenced by the degradation \cite{salcfsgan}, therefore, it is important to predict distortion-aware salient regions, which can also be used for guiding optimization direction. Therefore, we extend the visual saliency prediction task and define a distortion-aware saliency prediction task for AI generated omnidirectional images. Similar to the eye-tracking data analysis in regular saliency tasks, we first label the center of the distorted salient regions in an image, then the fixation maps are smoothed with a $0.4^{\circ}$ Gaussion kernel to obtain a continuous distortion-aware saliency map.
% Fig. \ref{fig:heatmap} presents two examples of the original images and corresponding distortion-aware saliency heatmaps, our saliency maps accurately highlight the salient distorted regions in the image.
\vspace{-12pt}
\subsection{Database Analysis}
\subsubsection{Statistics analysis for images in OHF2024 database}
\begin{figure*}[!t]
\vspace{-21pt}
\centering
\includegraphics[width=6.5in]{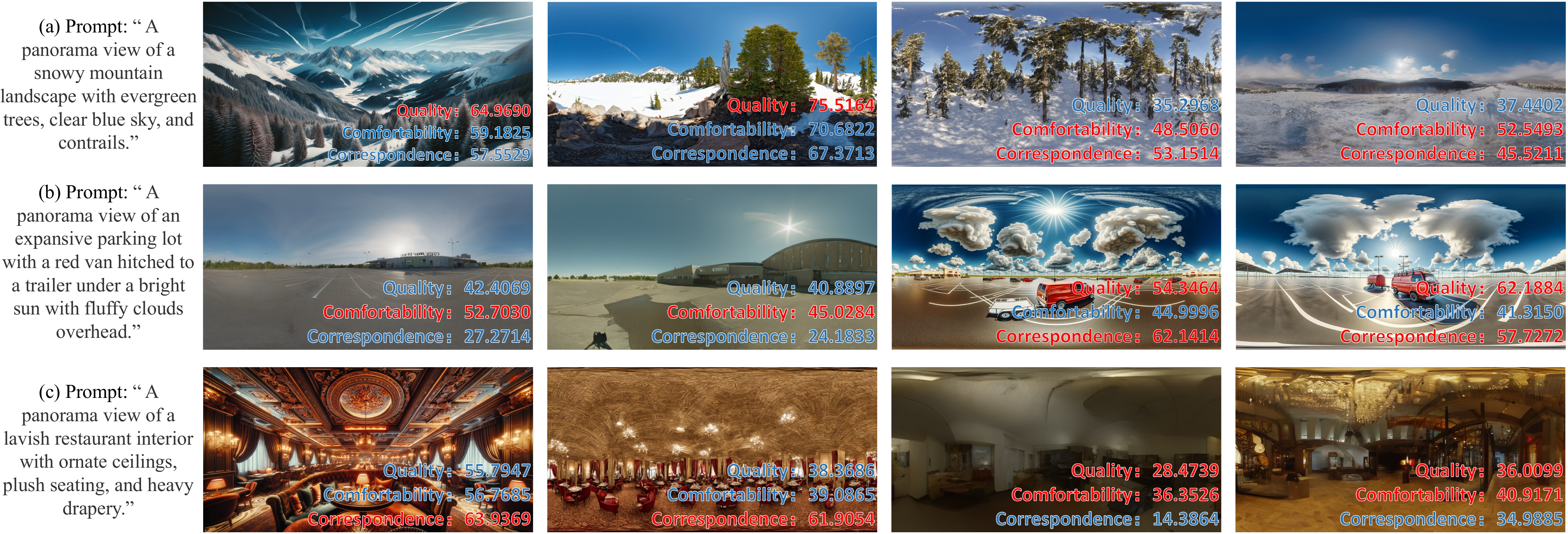}
\vspace{-12pt}
\caption{Comparison of the differences between three evaluation perspectives. (a) The left two omnidirectional images have better quality, but worse comfortability and correspondence, while the right two examples are contrary. (b) The left two omnidirectional images have better comfortability, but worse quality and correspondence, while the right two examples are contrary. (c) The left two omnidirectional images have better correspondence, but worse quality and comfortability, while the right two examples are contrary.}
\vspace{-15pt}
\label{fig:comparison}
\end{figure*}
We conduct statistical analysis for our OHF2024 database in terms of four low-level vision feature dimensions including: ``sharpness'', ``brightness'', ``colorfulness'' and ``contrast''. For simplicity, we refer to these four features as ${Ci}, (i = 1, 2, 3, 4)$, respectively. Some images from real-world omnidirectional databases, \textit{i.e.}, Matterport3D \cite{matterport} and SUN360 \cite{sun360} are used for comparison. Fig. \ref{fig:kernel} shows the kernel distribution of each feature for these three databases. It can be observed that the generated omnidirectional images have a wide distribution in ``sharpness'' and ``colorfulness'' features, showing their diversity. For the ``contrast'' feature, generated omnidirectional images show similar characteristics compared with other two natural omnidirectional databases.
However, for the ``brightness'' feature, AI-generated omnidirectional images show narrow distribution range compared to SUN360 \cite{sun360}.

To evaluate the distribution and uniformity of the databases over the four features, we further compute and compare the relative range and uniformity of coverage.
The relative range is calculate as:
\vspace{-3pt}
\begin{gather}
R_i^{d} = \frac{max(C_i^{d})-min(C_i^{d})}{max_d(C_i^d)}
\end{gather}
% \vspace{-3pt}
where $C_i^{d}$ refers to the data distribution of database $d$ on feature $i$, and $max_d(C_i^d)$ refers to the maximum value on feature $i$ across all databases.
Uniformity of coverage is calculated as the entropy of the B-bin histogram of $C_i^d$ over all sources for each database $d$:
\vspace{-3pt}
\begin{gather}
U_i^{d} = -\sum_{b=1}^{B} p_{b}log_{B}p_{b}
\end{gather}
% \vspace{-3pt}
where $p_b$ is the normalized number of souces in bin $b$ for each feature in each database.
\begin{figure}[!t]
% \vspace{-21pt}
\centering
\includegraphics[width=3.3in]{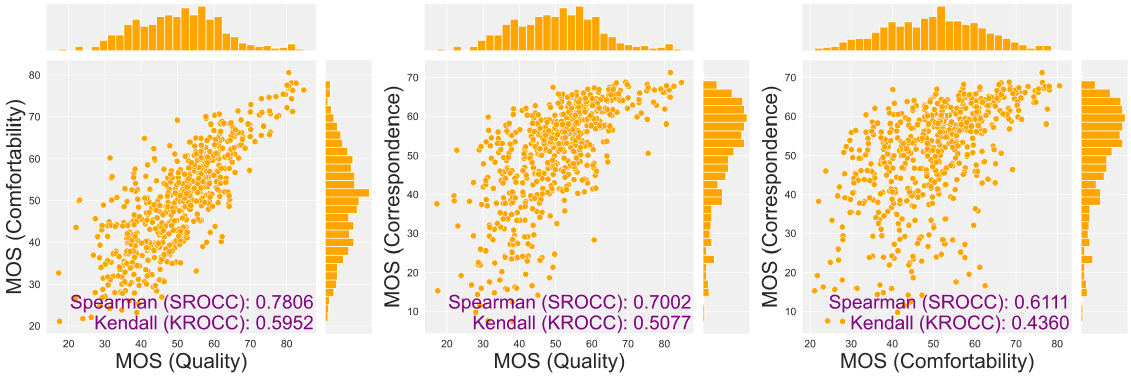}
\vspace{-12pt}
\caption{Illustration of the MOS correlation between any two dimensions.}
\vspace{-18pt}
\label{fig:cor}
\end{figure}
The relative range and uniformity of coverage are plotted in Fig. \ref{fig:UR}, quantifying intra- and inter-database differences. It can be concluded from the figures that the images in our OHF2024 dataset cover diverse ranges regarding the four selected features.

\subsubsection{Preference analysis from three perspectives}
In order to further emphasize the different focus of our three evaluation dimensions and verify the necessity of a single AI generated omnidirectional image from these three perspectives respectively, three sets of examples are given in Fig. \ref{fig:comparison}. The four images in each set are generated based on the same prompt.
For the left two images in each set, one dimension score is significantly higher than other two scores, while for the right two images, that score is significantly lower than the other two scores. In Fig. \ref{fig:comparison} (a), Fig. \ref{fig:comparison} (b), Fig. \ref{fig:comparison} (c), the left two images have higher quality scores, comfortability scores, correspondence scores, respectively, while the right two images have lower scores in these corresponding dimensions.
We conclude that different rating perspectives can reflect different human preferences, which are related but distinct.

We also analyze the correlation between any two dimensions in our OHF2024 database. As shown in Fig.~\ref{fig:cor}, quality and comfortability show some correlation, as low-level attributes can affect perceived comfort. In contrast, correspondence shows low correlation with the other two, further supporting the rationality of our three-perspective design.
Combining the analysis above, the assessment for a generated omnidirectional image should be performed from the three dimensions of quality, comfortability and correspondence dimensions, respectively.
\subsubsection{Relationship analysis between human visual preference and distortion-aware saliency}\label{relation}
\begin{figure}[!t]
\centering
\includegraphics[width=3.3in]{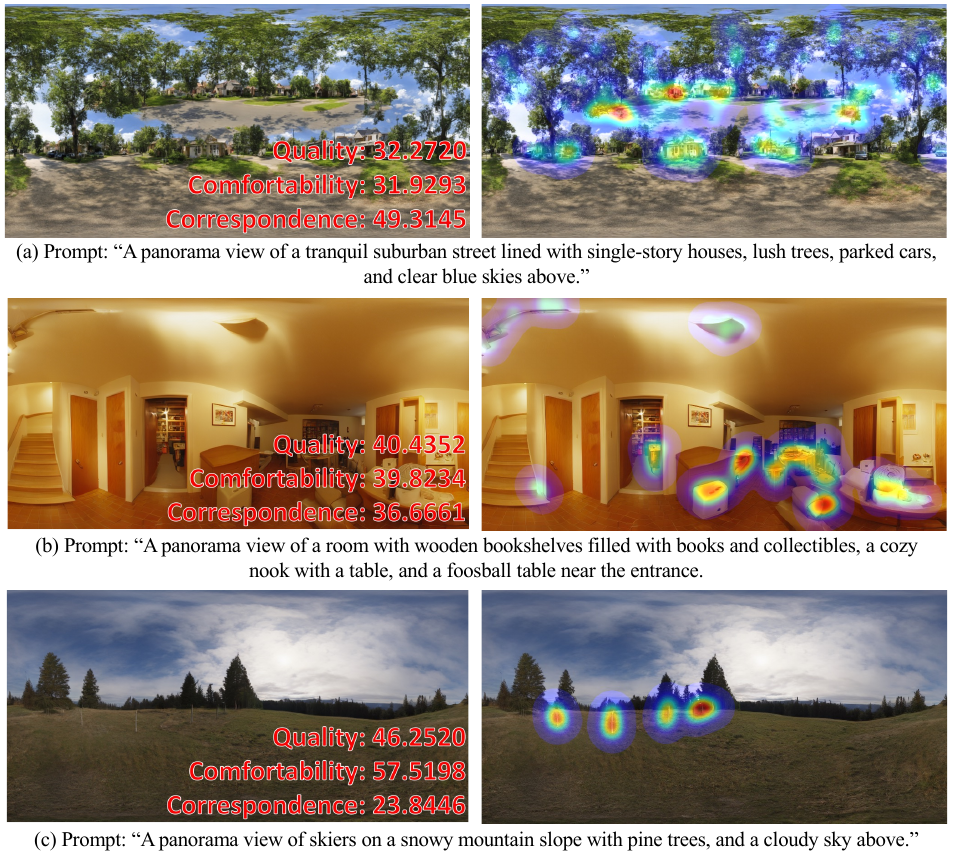}
\vspace{-9pt}
\caption{Examples of the human visual experience scores and distortion-aware saliency maps for AI-generated omnidirectional images. Left column: AIGODIs and the MOSs from our OHF2024 database. Right column: corresponding distortion-aware saliency maps.}
\vspace{-21pt}
\label{fig:analysis}
\end{figure}
Fig. \ref{fig:analysis} analyzes the relationship between human visual experience and distortion-aware saliency. The left column displays three AIGODIs from our OHF2024 database with the corresponding MOSs from three perspectives, respectively, and the right column shows the corresponding distortion-aware saliency maps. Comparing examples in Fig. \ref{fig:analysis} (a), (b) with Fig. \ref{fig:analysis} (c),  we observe that more distortion-aware salient regions correlate with significantly worse human visual experience in the quality and comfortability dimensions. Additionally, the correspondence dimension shows little correlation with the extent of distortion-aware saliency regions. For instance, in Fig. \ref{fig:analysis} (c), the MOS of the correspondence dimension is significantly lower than that of Fig. \ref{fig:analysis} (a) and (b), while the extent of the salient distorted region is significantly smaller due to its relatively high quality and comfortability.

However, it is evident that the location of distortion-aware salient regions correlates with the prompt, since our images are generated based on text prompts, and the text-corresponded regions are closely aligned with human visual attention. This finding highlights the importance of integrating text prompts into the text prompt in distortion-aware saliency prediction task for AI generated omnidirectional images.
\vspace{-6pt}
\section{Proposed Method}
\vspace{-3pt}
\begin{figure*}[!t]
\vspace{-21pt}
\centering
\includegraphics[width=7.0in]{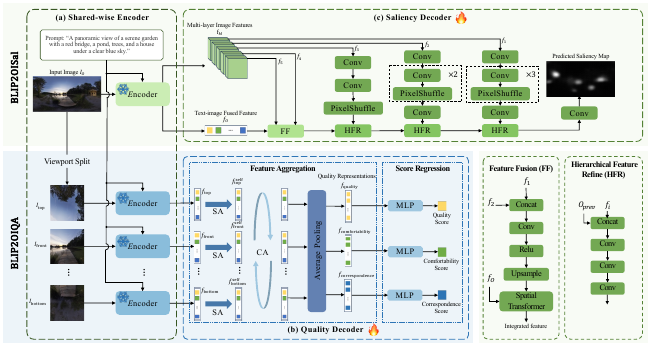}
\vspace{-12pt}
\caption{An Overview of the proposed BLIP2OIQA model (lower part) and BLIP2OISal model (upper part). (a) Both models utilize a shared-wise encoder to extract text-image fused feature from the input AIGODI and the prompt. (b) The BLIP2OIQA model employs a quality decoder, where the feature aggregation module integrates viewport-based text-image fused features and generate the final perspective-aware quality representations, which are then mapped to human visual experience scores through the score regression module.(c) The BLIP2OISal model is equipped with a saliency decoder, where the feature fusion (FF) module aggregates spatial information and text-image fused feature to form integrated feature, which is further refined with the multi-layer image features through three layers of hierarchical feature refine (HFR) module to produce the predicted distortion-aware saliency map.}
\vspace{-15pt}
\label{fig:blip2oiqa}
\end{figure*}
\begin{figure}[!t]
\centering
\includegraphics[width=3.3in]{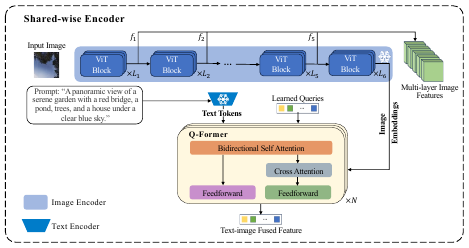}
\vspace{-9pt}
\caption{Detailed structure of the shared-wise encoder.}
\vspace{-18pt}
\label{fig:vl-backbone}
\end{figure}
% \vspace{-3pt}
% \vspace{-3pt}
In this section, to gain a comprehensive understanding of the quality of AI-generated omnidirectional images from the perspectives of human visual experience and distortion-aware saliency, we propose the BLIP2OIQA model and the BLIP2OISal model to achieve: (1) evaluating human visual experience for AI generated omnidirectional images from three perspectives including quality, comfortability and correspondence; (2) Distortion-aware saliency prediction for AI generated omnidirectional images. The model structure diagrams of BLIP2OIQA and BLIP2OISal are shown in Fig. \ref{fig:blip2oiqa}. Both models take images and the corresponding text prompts as inputs, with text-image fused feature obtained via the shared-wise encoder. Two well-designed decoders are adopted to accomplish different tasks. For the BLIP2OIQA model, we employ a quality decoder to generate the final quality scores, while for the BLIP2OISal model, we design a saliency decoder to predict distortion-aware saliency maps. The detailed architecture of the two proposed models are are given in section \ref{blip2oiqa} and \ref{BLIP2OISal}.
\vspace{-12pt}
\subsection{BLIP2OIQA}\label{blip2oiqa}
The structure of our BLIP2OIQA model is illustrated in the lower part of Fig. \ref{fig:blip2oiqa}. It consists of three main parts. (1) An AI-generated omnidirectional image is first segmented into six viewport images. Each viewport image is then processed through the shared-wise encoder with the text prompt to obtain viewport-based text-image fused features. (2) The feature aggregation module integrates the viewport-based text-image fused features to derive perspective-aware quality representations. (3) Quality representations are mapped to score spaces via the score regression module.
\subsubsection{Viewport-based feature extraction}
Given an AIGODI $I_O$ as input, six viewport images, denoted as $I_{\text{top}}$, $I_{\text{front}}$, $I_{\text{left}}$, $I_{\text{right}}$, $I_{\text{back}}$, $I_{\text{bottom}}$ are rendered with the field of view (FOV) set to $110^{\circ}$ as suggested in \cite{110}. These six viewpoints are perpendicular to each other and cover the entire visual content of the omnidirecional image. 

For each viewport, we use a shared-wise encoder to extract text-image fused features.
The shared encoder takes the viewport image and the corresponding text prompt as inputs to generate text-image fused feature, which is formulated as:
% \vspace{-3pt}
\begin{align}
f_{v} = E(I_{v}, T_{\text{prompt}})
\end{align}
% \vspace{-3pt}
where $v \in \{\text{top}, \text{front}, \text{left}, \text{back}, \text{right}, \text{bottom}\}$.
Detailed structure of the shared-wise encoder is shown in Fig. \ref{fig:vl-backbone}.
The input image is transformed into image embeddings through the image encoder while the input text prompt is encoded into text tokens through the text encoder. Both encoders are initialized with pretrained weights and are partially frozen to reduce computational costs as suggested in \cite{aigciqa_model}.
% In practice, we utilize the frozen visual encoder ViTG/14 from EVA-CLIP \cite{eva-clip} and the text encoder from BLIP \cite{blip}.
A multi-modal Q-Former is employed to generate viewport-based text-image fused feature by interacting with text and image features through a set of learnable queries, enabling visual and textual feature fusion.
Specifically, the queries interact with each other and the text tokens through the bidirectional self-attention layers, and interact with image embeddings through the cross attention layers.

Additionally, we extract five layers of image features from the transformer blocks of the image encoder to create multi-layer image features, which are utilized in the BLIP2OISal model.
\subsubsection{Attention-based feature aggregation}
We observe that human visual experience for AIGODIs is significantly influenced by viewport correspondence, particularly in the comfortability and correspondence dimensions. Specifically, for comfortability, users tend to pay attention to both intra-viewport degradations such as distortions and deformations, and inter-viewport consistency in styles and colors when browsing the whole image. For correspondence, users concentrate on the presence of the prompt objects in intra-viewport, and the relationship between the prompt objects across inter-viewport.
We leverage these observations to aggregate viewport-based text-image fused features, thereby comprehensively capturing visual experience features of AI generated omnidirectional images.

We employ self-attention and cross-attention modules to simulate the cognitive process of the users when browsing the AIGODIs in a 360-degree view, and then generate the final quality representations. For the extracted viewport-based text-image fused features $f_i$, $i \in \{\text{top}, \text{front}, ..., \text{bottom}\}$, we first apply the self-attention module to thoroughly capture intra-viewport quality information and generate viewport-based quality representations $f^{\text{self}}_i$. Next, we adopt the cross-attention modules for inter-viewport features, and generate three sets of perspective-aware quality representations related to quality, comfortability and correspondence perspectively. Specifically, each viewport-based quality representation $f^{\text{self}}_{i}$ interacts with others through cross-attention layers with shared weights, producing multiple sets of integrated features $f^{\text{cross}}_{(i,j)}$. These features are then averaged using an Average Pooling layer to obtain perspective-aware quality representations,
We employ three cross-attention layers with different weights to learn perspective-aware quality representations for the quality, comfortability, and correspondence dimensions, respectively, 
which are denoted as $f_{\text{quality}}$, $f_{\text{comfortability}}$ and $f_{\text{correspondence}}$. Our feature aggregation module can be described as:
\vspace{-3pt}
\begin{align}
&f^{\text{self}}_{i} = \text{SelfAttn}(f_i)\\
f^{\text{cross}}_{(i, j),k} &= \text{CrossAttn}_k(f^{\text{self}}_{i}, f^{\text{self}}_{j})\\
&\quad f_k = \frac{1}{N} \sum_{(i,j)} f^{\text{cross}}_{(i,j),k}
\vspace{-6pt}
\end{align}
% \vspace{-6pt}
where $i,j \in \{\text{top}, \text{front}, \text{left}, \text{back}, \text{right}, \text{bottom} \}$, $i\not= j$, $k \in \{ \text{quality}, \text{comfortability}, \text{correspondence}\}$.
\subsubsection{score regression}
We map the perspective-aware quality representations to the score space using three MLP layers as score regressors respectively. To reduce model complexity, the structure and loss function of the three regressors keep the same.
\vspace{-9pt}
\subsection{BLIP2OISal}\label{BLIP2OISal}
The structure of our BLIP2OISal model is illustrated in the upper part of Fig. \ref{fig:blip2oiqa}. The BLIP2OISal model is mainly composed of a shared encoder with the BLIP2OIQA and a saliency decoder. Given an AI-generated omnidirectional image and the corresponding text prompt, we first encode them into text-image fused features using the BLIP-2 model \cite{blip2}, and also extract multi-layer image features from the image encoder of BLIP-2. In the decoder part, the last two layers of the multi-layer image features are fed into the feature fusion (FF) module along with the text-image fused feature to integrate spatial information with text-image information. Subsequently, three hierarchical feature refine (HFR) modules are employed to progressively combine the integrated feature with layers from the multi-layer image features, each refined to an appropriate scale, ultimately producing the predicted distortion-aware saliency map.
\subsubsection{feature extraction}\label{featureextractoin}
In Section \ref{relation}, we have found that text prompts are strongly correlated with the distortion-aware saliency maps of AI generated omnidirectional images. Consequently, given an AI generated omnidirectional image $I_O$ and its text prompt as inputs, we use the shared-wise encoder which is same with the BLIP2OIQA model to integrate text-image fused feature. Following previous saliency prediction works, we directly use $I_O$ as the input image without the viewport split process to reduce computational complexity. The multi-layer image features extracted from the image encoder as shown in Fig. \ref{fig:vl-backbone} are also adopted here, which interact with the subsequent modules to preserve richer spatial information. We denote the multi-layer image features as:
\vspace{-6pt}
\begin{align}
    f_M = \{f_1, f_2, ..., f_5 \}
\vspace{-6pt}
\end{align}
% \vspace{-6pt}
In the subsequent modules, $f_i$ is first refined to the desired scale before interacting with the text-image fused feature in the FF and HFR modules as shown in Fig. \ref{fig:blip2oiqa}.
\subsubsection{feature fusion}
The text-image fused feature obtained in Section \ref{featureextractoin} does not effectively preserve spatial information, which is crucial for saliency prediction, thus, it is important to integrate the fused text-image features and multi-layer image spatial features to predict the distortion-aware saliency. As shown in Fig. \ref{fig:blip2oiqa}, the image features of the last two layers are first concatenated and then refined to desired scale before interacting with text-image fused feature through a spatial transformer, generating the integrated feature.
Three hierarchical feature refine modules (HFR) are applied to interact multi-layer image features $f_3$, $f_2$ and $f_1$ with the integrated feature at multiple scales, thus preserving spatial information while fully integrating with text features. Detailed structure of HFR module is also illustrated in Fig. \ref{fig:blip2oiqa}. 
\vspace{-9pt}
\subsection{Training Strategy}

\begin{table}
\vspace{-21pt}
\begin{center}
\caption{Performance comparision of the state-of-the-art NR-IQA models on Evaluating AI generated omnidirectional images from the perspectives of quality, comfortability and correspondence.
% $\diamondsuit$ and $\heartsuit$ denote scene-based and AIGC model-based train/test split for DNN-based IQA models, respectively.
The best performances are marked in \textcolor{red}{RED} and the second-best performances are marked in \textcolor{blue}{BLUE}.}
\vspace{-6pt}
\label{tab:BLIP2OIQA}
\scalebox{0.85}{
\renewcommand\arraystretch{0.9}
\begin{tabular}{l|cc|cc|cc}
\toprule
% Split & \multicolumn{9}{c|}{Scene-Based}& \multicolumn{9}{c}{Model-Based}\\
% \midrule
Dimension & \multicolumn{2}{c|}{Quality} & \multicolumn{2}{c|}{Comfortability} & \multicolumn{2}{c}{Correspondence}\\
\midrule
Model&SRCC&PLCC&SRCC&PLCC&SRCC&PLCC\\
\midrule
QAC~\cite{QAC}&0.2972&0.2483&0.3358&0.3048&0.2368&0.0710\\
BMPRI~\cite{bmpri}&0.5267&0.6575&0.3852&0.4989&0.3803
&0.3909\\
NIQE \cite{niqe}&0.7123&0.7150&0.5889&0.6684&0.6674&0.6088\\
ILNIQE \cite{ilniqe}&0.1797&0.2167&0.1675&0.2316&0.3291&	0.3553\\
HOSA \cite{hosa}&0.7469&0.7736&0.5160&0.5636&{0.7039}&	0.6496\\
BPRI-PSS \cite{pri}&0.4377&0.4883&0.2842&0.2657&0.5470&0.6247\\
BPRI-LSSs \cite{pri}&0.3757&0.5345&0.3088&0.4484&0.1944&0.2400\\
BPRI-LSSn \cite{pri}&0.4493&0.5906&0.2619&0.4019&0.5585&
0.5430\\
BPRI \cite{pri}&0.4285&0.5655&0.3497&0.4862&0.2581&0.2946\\
FISBLIM \cite{fisblim}&0.7036&0.6953&0.5167&0.4660&0.6618&0.5823\\
BRISQUE \cite{brisque}&0.1944&0.4109&0.1833&0.3371&0.2344&0.2784\\
\midrule
CLIPScore \cite{clipscore}&0.3353&0.3464&0.2165&0.2420&0.4150&0.5346\\
BLIPScore \cite{blip2}&0.3839&0.3509&0.2680&0.2542&0.5019&0.5912\\
\midrule
CNNIQA \cite{cnniqa}&0.7528&0.7248&0.6799&0.6769&0.5905&0.6835\\
Resnet18 \cite{resnet}&0.8237&0.8120&0.7503&0.7348&\textcolor{blue}{0.7865}&0.7770\\
Resnet34 \cite{resnet}&0.7439&0.7424&0.6272&0.5971&0.6509&0.7352 \\
VGG16 \cite{vgg}&0.8242&0.7783&0.7561&0.7132&0.7104&0.7436\\
VGG19 \cite{vgg}&0.8416&0.7960&0.7621&0.7333&0.7233&0.7189\\
HyperIQA \cite{hyperiqa}&{0.8710}&{0.8295}&0.7733&0.7833&0.7821&0.7755\\
MANIQA \cite{maniqa}&\textcolor{blue}{0.8793}&\textcolor{blue}{0.8606}&\textcolor{blue}{0.8130}&\textcolor{blue}{0.8345}&{0.7744}&\textcolor{blue}{0.8153} \\
TReS \cite{tres}&0.8231&0.8140&0.7556&0.7972& 0.7558&0.7508\\ \rowcolor{gray!20}
BLIP2OIQA (Ours) &\textcolor{red}{0.9074}&\textcolor{red}{0.8954}&\textcolor{red}{0.8763}&\textcolor{red}{0.9011}& \textcolor{red}{0.8354}&\textcolor{red}{0.8600}\\
\bottomrule
\end{tabular}
}
\vspace{-18pt}
\end{center}
\end{table}
As introduced in Section \ref{blip2oiqa} and Section \ref{BLIP2OISal}, for both BLIP2OISal and BLIP2OIQA models, we keep the image encoder and the text encoder partially frozen with the fix rate set to $0.7$, and keep the Q-Former in the shared-wise encoder frozen during training to save computational costs. As suggested in \cite{aigciqa_model}, we initialize the image encoder with ViTG/14 from EVA-CLIP \cite{eva-clip} and utilize the text encoder from BLIP \cite{blip}. And we initialize the Q-Former with a pretrained BLIP-2 model \cite{blip2}. In the following we describe the loss functions of BLIP2OIQA and BLIP2OISal in detail.
\subsubsection{BLIP2OIQA}
Following \cite{aigciqa_model}, we use L1 loss as the loss function for BLIP2OIQA, which can be formulated as follows:
\vspace{-9pt}
\begin{align}
    \mathcal{L} = \frac{1}{N} \sum_{i=1}^{N} |q_{\text{predict}}(i) - q_{\text{label}}(i)|
\end{align}
\vspace{0pt}
where $q_{\text{predict}}(i)$ is the predicted score for the $i^{th}$ perspective and $q_{\text{label}}(i)$ is the corresponding MOS derived from subjective experiment. $N$ is the total number of evaluation dimensions, which is $3$ for our OHF2024 database.
\subsubsection{BLIP2OISal}
We introduce a loss function formed by a linear combination of various evaluation metrics to obtain better distortion-aware saliency prediction results in terms of various evaluation metrics. We define the overall loss function as:
\vspace{-9pt}
\begin{align}
    \mathcal{L} = \alpha \cdot (1-\mathcal{L}_{\text{CC}}) + \beta \cdot (\mathcal{L}_{\text{KLD}})
\end{align}
where CC and KL are two widely used metrics for measuring the accuracy of the predicted saliency maps, while $\alpha$ and $\beta$ are two hyper-parameters that balance the two loss functions.

CC measures the linear relationship between the predicted saliency map $\hat{y}$ and ground truth $y$, which can be formulated as:
\vspace{-6pt}
\begin{align}
    CC = \frac{\sigma(\hat{y}, y)}{\sigma({\hat{y}})\cdot \sigma({{y})}}
\end{align}
\vspace{-6pt}

KLD adopts the dissimilarity of distributions between the predicted saliency map $\hat{P}$ and ground truth $P$, which can be computed as:
\vspace{-6pt}
\begin{align}
    KLD = \sum_{i=1}P\log \frac{P}{\hat{P}}
\end{align}
\vspace{-6pt}
where $\sum P=1$ and $\sum \hat{P}=1$
% \vspace{-6pt}

% \vspace{-3pt}
\section{Experiment}
\begin{figure*}[!t]
\vspace{-21pt}
\centering
\includegraphics[width=6.5in]{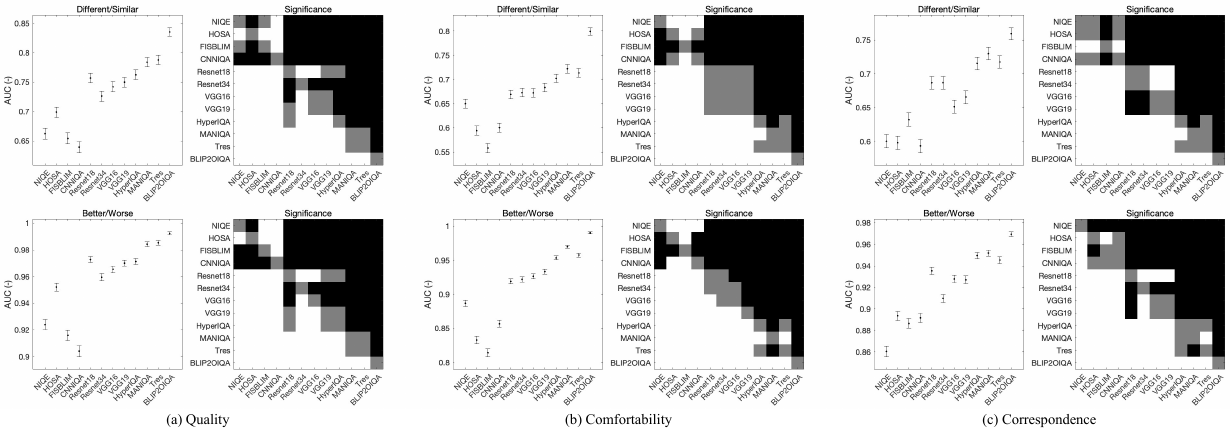}
\vspace{-12pt}
\caption{ROC analysis of 11 outstanding benchmark IQA models and the proposed BLIP2OIQA methods on the OHF2024 database from quality, comfortability and correspondence dimensions, respectively. The frist row illustrates the different vs. similar ROC analysis results, and the second row illustrates the better vs. worse analysis results. Note that the a white/black square in the significance figures means the row metric is statistically better/worse than the column one. A gray square means the row method and the column method are statistically indistinguishable.}
\vspace{-15pt}
\label{fig:auc}
\end{figure*}
In this section, we present a series of experiments to evaluate the performance of our proposed BLIP2OIQA and BLIP2OISal models. We first introduce the experiment protocol in detail. Then, we compare the performance of our BLIP2OIQA model with current the state-of-the-art IQA models in evaluating human visual experience for AI generated omnidirectional images. Then we evaluate the performance of our BLIP2OISal model in comparison with the state-of-the-art saliency prediction models on distortion-aware saliency prediction task. Finally, we conduct ablation studies to validate the effectiveness of our proposed components. 
\vspace{-9pt}
\subsection{Experiment Protocol}
The proposed OHF2024 database is used to validate the effectiveness of our models in evaluating human visual experience and distortion-aware saliency prediction for AI-generated omnidirectional images. 
In the experiments, the OHF2024 dataset is randomly split into training/testing sets with the ratio of 7:3 based on different scenes. For both BLIP2OIQA and BLIP2OISal models, the initial learning rate is set to $1e-05$ and decreased using the cosine annealing strategy. The adam optimizer with $\beta _1=0.9$ and $\beta _2=0.999$ is employed for training.

For the BLIP2OIQA model, we assess the performance by measuring the correlation between the predicted scores and the corresponding MOSs using Spearman Rank Correlation Coefficient (SRCC), Pearson Linear Correlation Coefficient (PLCC), and Kendall's Rank Correlation Coefficient (KRCC).
We also adopt receiver operating characteristic (ROC) analysis \cite{AUC1, AUC2} as an additional evaluation method for IQA models. Specifically, the ``Different \textit{vs.} Similar'' and ``Better \textit{vs.} Worse'' ROC analyses evaluate a model’s ability to distinguish between perceptually distinct or ranked image pairs, respectively.
% The receiver operation characteristic (ROC) analysis methodology \cite{AUC1, AUC2} is also adopted as an additional evaluation method for IQA metrics, which assesses two key aspects: \textit{i.e.}, whether two samples distincly differ in visual experience and if they do, which one is superior. The ``Different \textit{vs.} Similar ROC Analysis'' determines whether various objective metrics can distinguish image pairs with and without significant differences, while the ``Better \textit{vs.} Worse ROC Analysis'' is applied to test if various objective metrics can distinguish images with positive and negative differences. The area under the ROC curve (AUC) values of two analysis is reported in this paper, where higher values indicate better performance.

 To measure the performance of the BLIP2OISal model on the distortion-aware saliency prediction task, we choose five commonly used consistency metrics including location-based metrics such as AUC and NSS, and distribution-based metrics including CC, SIM, and KLD, with the reference to prior works \cite{transalnet, ref4}.
\vspace{-9pt}
\subsection{Performance on Evaluating Human Visual Preference}
We compare the performance of our proposed BLIP2OIQA model with 21 state-of-the-art NR-IQA models. For handcrafted-based models including QAC \cite{QAC}, BMPRI \cite{bmpri}, NIQE \cite{niqe}, ILNIQE \cite{ilniqe}, HOSA \cite{hosa}, FISBLIM \cite{fisblim}, BPRI-PSS \cite{pri}, BPRI-LSSs \cite{pri}, BPRI-LSSn \cite{pri}, BPRI \cite{pri} and BRISQUE \cite{brisque}, we directly employ them to predict the preference scores for the omnidirectional images in our database. For vision-language pretrained models including CLIPScore \cite{clipscore} and BLIPScore \cite{blip}, we calculate the scores using the cosine similarity between the text and image embeddings. For deep learning-based models including CNNIQA \cite{cnniqa}, Resnet18 \cite{resnet}, Resnet34 \cite{resnet}, VGG16 \cite{vgg}, VGG19 \cite{vgg},  HyperIQA \cite{hyperiqa}, MANIQA \cite{maniqa} and TReS \cite{tres}, we retrain them on OHF2024 using the official training settings.

% \begin{enumerate}
% \item[$\bullet$]\textbf{Handcrafted-based} models, which include QAC \cite{QAC}, BMPRI \cite{bmpri}, NIQE \cite{niqe}, ILNIQE \cite{ilniqe}, HOSA \cite{hosa}, FISBLIM \cite{fisblim}, BPRI-PSS \cite{pri}, BPRI-LSSs \cite{pri}, BPRI-LSSn \cite{pri}, BPRI \cite{pri} and BRISQUE \cite{brisque}.
% \item[$\bullet$]\textbf{Vision-language pretrained} models, including CLIPScore \cite{clipscore} and BLIPScore \cite{blip}.
% \item[$\bullet$]\textbf{Deep learning-based} models, including CNNIQA \cite{cnniqa}, Resnet18 \cite{resnet}, Resnet34 \cite{resnet}, VGG16 \cite{vgg}, VGG19 \cite{vgg},  HyperIQA \cite{hyperiqa}, MANIQA \cite{maniqa} and TReS \cite{tres}.
% \end{enumerate}
 \begin{table}
% \vspace{-21pt}
\begin{center}
\scriptsize
\caption{Performance comparison of the state-of-the-art NR-IQA models with train/test splitting according to AIGC models. The best performances are marked in \textcolor{red}{RED} and the second-best performances are marked in \textcolor{blue}{BLUE}.}
\vspace{-6pt}
\label{tab:model-split}
\scalebox{0.95}{
\renewcommand\arraystretch{0.9}
\begin{tabular}{ l |c c|c c|c c}
\toprule
Dimension & \multicolumn{2}{c|}{Quality} & \multicolumn{2}{c|}{Comfortability} & \multicolumn{2}{c}{Correspondence}\\
\midrule
Model&SRCC&PLCC&SRCC&PLCC&SRCC&PLCC\\
\midrule
HyperIQA~\cite{hyperiqa}&{0.7293}&0.7281&\textcolor{blue}{0.7632}&0.8030&0.7640&0.7610\\
MANIQA~\cite{maniqa}&\textcolor{blue}{0.7630}&\textcolor{blue}{0.7431}&0.7500&\textcolor{blue}{0.8084}&0.7487&0.7389\\
TReS~\cite{tres}&0.7491&0.7416&0.7524&0.7914&\textcolor{blue}{0.7962}&\textcolor{blue}{0.7900}\\ \rowcolor{gray!20}
BLIP2OIQA (Ours)&\textcolor{red}{0.7756}&\textcolor{red}{0.7509}&\textcolor{red}{0.7982}& \textcolor{red}{0.8238}&\textcolor{red}{0.8182}&\textcolor{red}{0.8190}\\
\bottomrule
\end{tabular}}
\vspace{-6pt}
\end{center}
\vspace{-18pt}
\end{table}
% For handcrafted-based models, we directly employ them to predict the preference scores for the omnidirectional images in our database. For CLIPScore \cite{clipscore} and BLIPScore \cite{blip}, we calculate the scores using the cosine similarity between the text and image embeddings. For deep learning-based IQA models, the training parameter settings are the same as the officially released code.

Table \ref{tab:BLIP2OIQA} demonstrate the performance comparison results. 
It can be observed that hand-crafted models show poor performance for evaluating the human visual experience of AIGODIs, and these models perform even worse for the dimensions of comfortability and text-image correspondence.
The current state-of-the-art deep learning-based models generally outperform hand-crafted models, but their performance is still not entirely satisfactory. Particularly, they struggle to achieve overall good performance in quality, comfortability and correspondence.
These models generally perform better in the quality dimension but worse in the comfortablity and correspondence dimensions, due to the unawareness of the authenticity and comfortable textures, as well as the ignoring of utilizing the text information.
The proposed BLIP2OIQA model shows state-of-the-art performance on all three perspectives, and achieves significant improvements over current state-of-the-art models in both comfortability and correspondence dimensions. This confirms the effectiveness of our model in evaluating human visual experience for AI generated omnidirectional images from multiple perspectives.
Fig.\ref{fig:auc} illustrates the AUC performance of the proposed BLIP2OIQA and other 11 outstanding benchmark IQA methods, which further demonstrates that our BLIP2OIQA model significantly outperforms other benchmarks methods on Different \textit{vs.} Similar and Better \textit{vs.} Worse Analysis on all three evaluation dimensions.
% \vspace{-9pt}
% \subsection{Cross-model Evaluation}

 As different ODI generative models exhibit distinct generation styles, it is essential to assess whether an IQA model can generalize across them. To this end, we also conduct cross-model evaluation, \textit{i.e.}, train/test split based on AIGC models, to validate the models' generalizability across different generative models. The results are shown in Table \ref{tab:model-split}. Our BLIP2OIQA model achieves the best performance, particularly on the ``correspondence'' dimension, showing its adaptability in evaluating human visual perception of ODIs generated by diverse AIGC models.
 \begin{table}
\vspace{-3pt}
\begin{center}
\vspace{-3pt}
\caption{Performance comparision of the state-of-the-art Saliency models on distortion-aware saliency prediction for AI generated omnidirectional images. The best performances are marked in \textcolor{red}{RED} and the second-best performances are marked in \textcolor{blue}{BLUE}.}
\vspace{-6pt}
\label{tab:BLIP2OISal}
\scalebox{0.9}{
\renewcommand\arraystretch{0.9}
\begin{tabular}{ l |c|c|c|c|c}
\toprule
Model$\backslash$Metric & AUC$\uparrow$ & NSS$\uparrow$ & CC$\uparrow$ & SIM$\uparrow$ & KLD$\downarrow$\\
\midrule
AIM \cite{aim}&0.7994 &1.1000&0.2510 &0.2183 &9.3104\\
CA \cite{ca} &0.7692&1.0525&0.2588&0.2305&8.9496\\
CovSal \cite{covsal}&0.6215&0.5784&	0.1442&0.1991&10.094\\
GBVS \cite{gbvs}&0.7297&0.8749&0.2240&0.2133&9.2293\\
HFT \cite{hft}&0.7380&	0.9403&0.2373&0.2321&8.8655\\
IT \cite{it}&0.5469&0.3091&0.0747&0.1217&18.110\\
Judd \cite{judd}&0.7499&0.9163&	0.2269&	0.1836&	9.7636\\
Murray \cite{murray}&0.7835&1.0403&0.2544&0.2057&9.3753\\
PFT \cite{pft}&0.7443&0.9302&0.2327&0.2225&9.0721\\
SMVJ \cite{smvj}&0.7343&0.9035&0.2298&0.2140&9.2145\\
SR \cite{sr}&0.7399&0.8878&0.2295&0.2190&9.1276\\
SUN \cite{sun}&0.7891&1.2064&0.2562&0.2229&9.1283\\
SWD \cite{swd}&0.7294&0.9063&0.2213&0.2009&9.4484\\
\midrule
SALICON \cite{salicon1}&0.7119&1.6080&0.3146&0.2941&8.2621\\
TranSalNet-ResNet \cite{transalnet}&0.8073&1.2615&0.3035&0.2181&9.4058\\
TranSalNet-Dense \cite{transalnet}&\textcolor{blue}{0.8178}&1.3428&0.3231&0.2397&8.9139\\
Sal-CFS-GAN \cite{salcfsgan}&0.7623&1.1233&0.2571&0.2276&10.471\\
SAM-VGG \cite{sam}&0.7915&1.1839&0.2933&0.2442&8.9760\\
SAM-ResNet \cite{sam}&0.7746&1.1237&0.2761&0.2273&9.3306\\
VQSal \cite{ref4}&0.8002&\textcolor{blue}{1.6566}&\textcolor{blue}{0.3661}&\textcolor{blue}{0.2967}&\textcolor{blue}{8.1803}\\ \rowcolor{gray!20}
BLIP2OISal (Ours)&\textcolor{red}{0.8312}&\textcolor{red}{1.8867}&\textcolor{red}{0.4324}&\textcolor{red}{0.3204}&\textcolor{red}{7.2766}\\
\bottomrule
\end{tabular}
}
\vspace{-18pt}
\end{center}
\vspace{-6pt}
\end{table}
 \vspace{-12pt}
\subsection{Performance on Distortion-aware Saliency Prediction}
We compare our proposed BLIP2OISal model with 20 current state-of-the-art saliency prediction models to validate the effectiveness of our model on distortion-aware saliency prediction task. For traditional models including AIM \cite{aim}, CA \cite{ca}, CovSal \cite{covsal}, GBVS \cite{gbvs}, HFT \cite{hft}, IT \cite{it}, Judd \cite{judd}, Murray \cite{murray}, PFT \cite{pft}, SMVJ \cite{smvj}, SR \cite{sr}, SUN \cite{sun} and SWD \cite{swd}, we directly use them to predict saliency maps for the given AIGODIs. For deep learning-based models including SALICON \cite{salicon1}, TranSalNet \cite{transalnet}, Sal-CFS-GAN \cite{salcfsgan} and VQSal \cite{ref4}, we retrain them on the proposed OHF2024 database using their officially realeased code and utilizing the same training/testing sets as those of BLIP2OISal.

As shown in Table \ref{tab:BLIP2OISal}, nearly all state-of-the-art saliency prediction models perform unsatisfactorily on the distortion-aware saliency prediction task, particularly in terms of CC and NSS metrics. This is due to the fact that these models are primarily designed to identify general salient regions, whereas distortion-aware salient regions have unique characteristic.
Specifically, distortion-aware salient regions address both visually salient and the low-quality parts of an image, which is significantly different from traditional saliency definition. Moreover, distortion-aware saliency prediction is also correlated with the text prompt, as discussed in Section \ref{relation}. Therefore, conventional saliency prediction models lack sufficient adaptability for this specialized task. Although deep learning-based models perform better than classical models due to their adaption ability during training, their performance is still far from satisfactory.
Finally, from the quantitative perspective, our proposed BLIP2OISal model achieves state-of-the-art performance on distortion-aware saliency prediction for AIGODIs, which can predict distortion-aware saliency more effectively.
\vspace{-12pt}
\subsection{Ablation Studies}
\vspace{-3pt}

\begin{table*}
\vspace{-21pt}
\begin{center}
\caption{Ablation studies on the architecture of BLIP2OIQA. The best performances are marked in \textcolor{red}{RED} and the second-best performances are marked in \textcolor{blue}{BLUE}.}
\vspace{-6pt}
\label{tab:ablation1}
\renewcommand\arraystretch{0.9}
\begin{tabular}{ l |c c c|c c c|c c c}
\toprule
Dimension & \multicolumn{3}{c|}{Quality} & \multicolumn{3}{c|}{Comfortability} & \multicolumn{3}{c}{Correspondence}\\
\midrule
Methods&SRCC&KRCC&PLCC&SRCC&KRCC&PLCC&SRCC&KRCC&PLCC\\
\midrule
w/o viewports&0.8934&0.7081&0.8790&0.8385&0.6583&0.8682&0.8055&0.6105&0.8396\\
w/o sa w/o ca&0.8899&0.7055&0.8786&0.8395&0.6564&0.8690&0.8146&0.6208&0.8576\\
w/o ca&0.8980&0.7151&0.8844&0.8563&0.6777&0.8851&0.8288&0.6306&0.8330  \\
w/o sa&\textcolor{red}{0.9079}&\textcolor{red}{0.7317}&\textcolor{red}{0.8993}&\textcolor{blue}{0.8714}&\textcolor{blue}{0.6993}&\textcolor{blue}{0.8959}&\textcolor{blue}{0.8327}&\textcolor{blue}{0.6443}&\textcolor{blue}{0.8551} \\
\rowcolor{gray!20}
BLIP2OIQA &\textcolor{blue}{0.9074}&\textcolor{blue}{0.7272}&\textcolor{blue}{0.8954}&\textcolor{red}{0.8763}&\textcolor{red}{0.7036}&\textcolor{red}{0.9011}& \textcolor{red}{0.8354}&\textcolor{red}{0.6417}&\textcolor{red}{0.8600}\\
\bottomrule
\end{tabular}
\vspace{-18pt}
\end{center}
\end{table*}
\begin{table}
\vspace{-6pt}
\begin{center}
\caption{The exploration of integrating text features.  The best performances are marked in \textcolor{red}{RED} and the second-best performances are marked in \textcolor{blue}{BLUE}.}
\vspace{-6pt}
\label{tab:ablation2}
\scalebox{0.9}{
\renewcommand\arraystretch{0.9}
\begin{tabular}{l|c|c|c|c|c}
\toprule
 Method$\backslash$Metric & AUC$\uparrow$ & NSS$\uparrow$ & CC$\uparrow$ & SIM$\uparrow$ & KLD$\downarrow$\\
\midrule
Type \uppercase\expandafter{\romannumeral1}&0.8251&	1.7745&0.4152&0.3169&7.3454  \\ 
Type \uppercase\expandafter{\romannumeral2}&\textcolor{blue}{0.8358}&\textcolor{blue}{1.8218}&\textcolor{blue}{0.4283}&0.3137&7.3885  \\
Type \uppercase\expandafter{\romannumeral3}&\textcolor{red}{0.8386}&1.7894&0.4209&0.3188&7.3390 \\
Type \uppercase\expandafter{\romannumeral4}&0.8281&1.7966&0.4201&\textcolor{blue}{0.3182}&\textcolor{blue}{7.3380}\\
\rowcolor{gray!20}
BLIP2OISal&{0.8312}&\textcolor{red}{1.8867}&\textcolor{red}{0.4324}&\textcolor{red}{0.3204}&\textcolor{red}{7.2766}\\
\bottomrule
\end{tabular}
}
\end{center}
\vspace{-21pt}
\end{table}
We conduct ablation experiments on both the BLIP2OIQA and BLIP2OISal models to validate the effectiveness of our proposed components. All experiments are performed on the OHF2024 database with consistent data partitioning and parameter settings.

\subsubsection{Ablation study for BLIP2OIQA}

We primarily validate the effectiveness of our proposed viewport-based feature extraction and feature aggregation modules. The results are presented in Table \ref{tab:ablation1}. ``w/o viewports'' indicates directely inputting the AI generated omnidirectional image and its prompt into the shared-wise encoder without the viewport-split process and map the outputs into scores using three MLP layers \cite{aigciqa_model}. ``w/o sa w/o ca'' indicates removing both the self-attention and cross-attention module, \textit{i.e.}, the viewport-based text-image fused features are simply integrated through Average Pooling layer before mapped to human visual experience scores. ``w/o sa'' and ``w/o ca'' represent setups where only the self-attention layers or only the cross-attention modules are removed, respectively.
\begin{table}
\vspace{-6pt}
\begin{center}
\caption{The exploration of integrating spatial features. The best performances are marked in \textcolor{red}{RED} and the second-best performances are marked in \textcolor{blue}{BLUE}.}
\vspace{-9pt}
\label{tab:ablation3}
\scalebox{0.9}{
\renewcommand\arraystretch{0.9}
\begin{tabular}{l|c|c|c|c|c}
\toprule
 Method$\backslash$Metric & AUC$\uparrow$ & NSS$\uparrow$ & CC$\uparrow$ & SIM$\uparrow$ & KLD$\downarrow$\\
\midrule
Type \uppercase\expandafter{\romannumeral1}&0.7331&1.2804&0.3015&0.1638&10.182\\ 
Type \uppercase\expandafter{\romannumeral2}&0.7776&1.2803&0.3066&0.1687&10.073\\
Type \uppercase\expandafter{\romannumeral3}&\textcolor{blue}{0.8409}&1.8011&0.4241&0.3089&7.5137 \\
Type \uppercase\expandafter{\romannumeral4}&\textcolor{red}{0.8495}&\textcolor{blue}{1.8222}&\textcolor{blue}{0.4277}&\textcolor{blue}{0.3092}&\textcolor{blue}{7.4924}\\
\rowcolor{gray!20}
BLIP2OISal&{0.8312}&\textcolor{red}{1.8867}&\textcolor{red}{0.4324}&\textcolor{red}{0.3204}&\textcolor{red}{7.2766}\\
\bottomrule
\end{tabular}
}
\vspace{-21pt}
\end{center}
\end{table}
When directly use AIGODIs and the corresponding text prompts as inputs, the model already outperforms state-of-the-art IQA models, confirming the advantage of employing the shared encoder to extract features for assessing human visual experience for AI generated omnidirectional images. As shown in Table \ref{tab:ablation1}, removing either module results in a performance drop. Notably, the cross-attention module significantly influences model performance across all three evaluation perspectives. Additionally, incorporating both self-attention and cross-attention modules yields further improvements overall, validating the effectiveness of our proposed attention-based feature aggregation framework.
\vspace{3pt}
\begin{figure*}[!t]
\vspace{-18pt}
\centering
\includegraphics[width=7.0in]{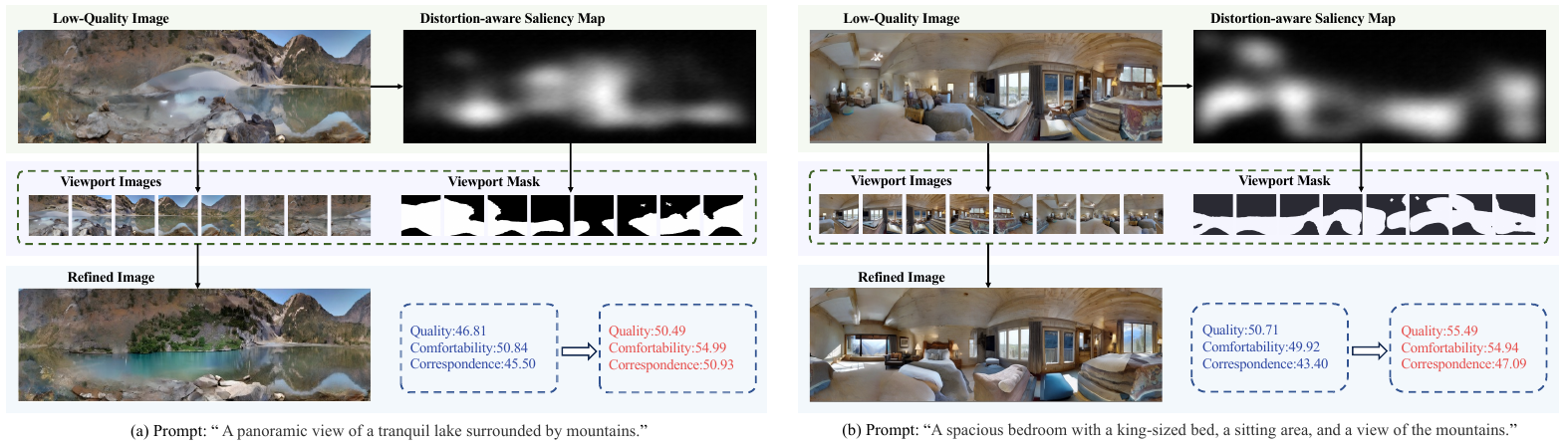}
\vspace{-12pt}
\caption{Two sets of examples to illustrate the workflow and results of our optimization process. The examples include the original low-quality image and its corresponding distortion-aware saliency map, the viewport images and the corresponding viewport masks (with white representing masked regions and black indicating non-masked regions), and the refined images after our optimization process. The refined images show performance improvement across all three evaluation dimensions compared to the original images.}
\vspace{-15pt}
\label{fig:inpainting}
\end{figure*}
\subsubsection{Ablation study for BLIP2OISal}
For BLIP2OISal, we propose using the feature fusion (FF) module and hierarchical feature refine (HFR) modules to integrate text features while preserving spatial information. Ablation experiments are conducted to validate the effectiveness of these two modules separately.

We first demonstrate the rationality of adopting the FF module to integrate text features, with results shown in Table \ref{tab:ablation2}.
We first replace the text-image fused features by the text tokens generated by the text encoder in the shared-wise encoder as input to the FF module, denoted as ``Type \uppercase\expandafter{\romannumeral1}''. Our saliency decoder utilizes one layer of FF module to aggregate spatial and text-image fused features. We also conduct ablation experiments to validate the impact of injecting text-image fused features at different layers. In the ``Type \uppercase\expandafter{\romannumeral2}'' version, the text-image fused feature is removed entirely, \textit{i.e.}, using only four layers of the HFR module as the saliency decoder. In the ``Type \uppercase\expandafter{\romannumeral3}'' version and ``Type \uppercase\expandafter{\romannumeral4}'' version, the text-image fused feature is injected into two and three layers, respectively.
% The diagrams of Type \uppercase\expandafter{\romannumeral1}, \uppercase\expandafter{\romannumeral2}, and \uppercase\expandafter{\romannumeral3} are illustrated in Figure \ref{fig:ablation2}.

The results indicate that using the text-image fused features generated by the shared encoder in the FF module significantly improves model performance compared to directly inputting text tokens from the text encoder. Additionally, incorporating the text-image fused feature in a single layer of FF module enhances the utilization of text information and improves distortion-aware saliency prediction accuracy. However, adding multiple layers of text-image fused features can lead to overfitting on our OHF2024 database.

We also conduct ablation experiments to validate the effectiveness of using multi-layer image features to preserve spatial information of images. The results are shown in Table \ref{tab:ablation3}.
When removing the multi-layer image features from the HFR module and only input single-layer or multi-layer image features into the FF module (Type \uppercase\expandafter{\romannumeral1} and Type \uppercase\expandafter{\romannumeral2}), the performance of the model significantly declines. This indicates that integrating image features from the image encoder into the HFR and FF modules greatly enhances the model's capacity to recover spatial information, thereby improving the performance on distortion-aware saliency prediction. The performance of the model gradually improves when using a single image feature layer as input to both the HFR and FF modules (Type \uppercase\expandafter{\romannumeral3}) and using three feature layers separately as inputs to the HFR and FF modules (Type \uppercase\expandafter{\romannumeral4}), which proves that progressively integrating multiple feature layers from the image encoder into the FF and HFR modules contributes to the enhanced model performance.
\vspace{-6pt}
\section{Feedback for Optimizing AI Generated Omnidirectional Images}
In this section, we demonstrate that our BLIP2OIQA and BLIP2OISal models can provide rich human visual information on the quality of AI-generated omnidirectional images, which can be utilized in an automatic optimization process to enhance the visual quality of AIGODIs. Given the limited number of existing AIGODI models, for the effective experiment validation, we choose MVDiffusion \cite{mvdiffusion} as our target model and partially modify the algorithms without retraining to achieve our demonstration objectives. The workflow illustrating how our BLIP2OIQA and BLIP2OISal models help in predicting human feedback and promote the optimization process is shown in Fig. \ref{fig:workflow}.

\vspace{-9pt}
\subsection{Rich Feedback Generation using BLIP2OIQA and BLIP2OISal}\label{sec:select}
We first demonstrate how our BLIP2OIQA model can help in evaluating human visual experience and thereby be used to filter AIGODIs. Following the procedure described in Section \ref{sec:collection}, we first re-generate 100 prompts describing 50 indoor and 50 outdoor scenes respectively, which cover a wide variety of scenarios. Using MVDiffusion \cite{mvdiffusion}, we then generate two omnidirectional images for each prompt, introducing diversity by randomly adjusting the generation parameters.

Next, we apply our BLIP2OIQA model to predict human visual experience for each image from three perspectives of quality, comfortability and correspondence. We then filter the images based on the average score across these dimensions. If both images of the same scene have average scores within an satisfactory range, they are directly outputted. If an image has a significantly unsatisfactory averaged score, we filter it out and proceed to make local adjustments based on the distortion-aware salient regions predicted by our BLIP2OISal model, which will be detailed in Section \ref{sec:modify}.
\vspace{-9pt}
\subsection{Optimization with Rich Feedback}\label{sec:modify}
% \begin{table}
% % \vspace{-6pt}
% \begin{center}
% \caption{BLIP2OIQA score and human annotation score for low-quality and high-quality AI generated omnidirectional images, before and after salient-distorted-region inpainting based on the BLIP2OISal algorithm.}
% \vspace{-6pt}
% \label{tab:userstudy}
% \scalebox{0.8}{
% \renewcommand\arraystretch{0.9}
% \begin{tabular}{l|l|cc|cc|cc}
% \toprule
%  &\centering{Dimension}&\makebox[1.7cm]{Quality}&\makebox[1.7cm]{Comfortability}& \makebox[1.7cm]{Correspondence}\\
% \midrule
% \multirow{2}{*}{LQ}&\centering{Before}&46.92&44.98&47.64\\ 
% &\centering{After}&49.07&47.69&50.07\\
% \midrule
% \multirow{2}{*}{HQ}&\centering{Before}&47.48&46.21&48.58\\ 
% &\centering{After}&49.48&48.74&50.52\\
% \bottomrule
% \end{tabular}
% }
% \vspace{-18pt}
% \end{center}
% \end{table}

\begin{table}

\centering

\caption{BLIP2OIQA score and human annotation score for low-quality and high-quality AI generated omnidirectional images, before and after salient-distorted-region inpainting based on the BLIP2OISal model.}
\vspace{-2mm} 
\centering
\resizebox{\linewidth}{!}{\begin{tabular}{lccccccc}
   \toprule
&Dimension& \multicolumn{2}{c}{Quality} & \multicolumn{2}{c}{Comfortability}  & \multicolumn{2}{c}{Correspondence}\\  \cmidrule(lr){3-4} \cmidrule(lr){5-6} \cmidrule(lr){7-8}
&Type/Metric& Human&BLIP2OIQA&Human&BLIP2OIQA& Human&BLIP2OIQA\\  
     \midrule
   \multirow{2}{*}{LQ} &\centering{Before}&46.31&46.92&43.89&44.98&46.62&47.64\\
    &\centering{After} &\textbf{50.66}&\textbf{49.07}&\textbf{45.83}&\textbf{47.69}&\textbf{51.24}&\textbf{50.07}\\
    \midrule
   \multirow{2}{*}{HQ} &\centering{Before}& {49.40}&{47.48}& {46.71}&{46.21}&{49.13}&{48.58}  \\
   &\centering{After} &\textbf{51.55}&\textbf{49.48}&\textbf{48.26}&\textbf{48.74}&\textbf{52.73}&\textbf{50.52}\\
    \bottomrule
  \end{tabular}
  } 
  \vspace{-2em}
  \label{tab:inpainting}
\end{table}

We slightly modify the MVDiffusion \cite{mvdiffusion} model to meet our requirements. Specifically, we divide the image into eight viewports as needed for the input format, and simultaneously split the distortion-aware saliency map predicted by BLIP2OISal into the corresponding eight viewports. For the need of binary input mask of the inpainting model, we set a threshold value $x_0$. For each viewport-based saliency map, if the single-channel pixel value is greater than $x_0$, the region is classified as a masked region that requires inpainting; otherwise, it is considered as a non-masked region. This can be expressed by the following equation:
\vspace{-3pt}
\begin{align}
M(i,j)\begin{cases}
1 & \text{if } S(i,j) > x_0 \\
0 & \text{if } S(i,j) \leq x_0
\end{cases}
\end{align}
where \( M(i,j) \) represents the mask value at pixel \((i,j)\) and \( S(i,j) \) represents the single-channel saliency value at the same pixel.

As illustrated in Fig. \ref{fig:inpainting}, we split the unsatisfied omnidirectional image and its corresponding distortion-aware saliency map into eight viewports. For each viewport, a mask is generated to identify the distorted regions that require to be optimized. Utilizing MVDiffusion \cite{mvdiffusion}, we then perform inpainting on these distorted regions within the viewports, generating a refined omnidirectional image. The refined image retains most of the original structure while effectively optimizing the visually unsatisfactory areas, resulting in a notable improvement in overall image quality, in terms of both visual perception and according to the evaluation metrics of BLIP2OIQA.

For the two omnidirectional images corresponding to each prompt generated in Section \ref{sec:select}, we classify them into low-quality (LQ) and high-quality (HQ) groups based on human experience scores predicted by BLIP2OIQA model. Then, we adopt the aforementioned optimization process for the images and record the average BLIP2OIQA scores of the refined images. A subjective study in also conducted as described in Section \ref{sec:subjective-experiment} to obtain human ratings. As shown in Table \ref{tab:inpainting}, optimization based on the distortion-aware saliency prediction leads to a significant improvement in both the human ratings and the BLIP2OIQA scores from all three perspectives, not only for the low-quality images but also for the high-quality set. This confirms the effectiveness of our proposed optimization process based on our BLIP2OISal model in enhancing the visual quality of AI generated omnidirectional images.
\vspace{-6pt}
\section{Conclusion}
This work aims to comprehensively study the quality issues of AI generated omnidirectional images and promote an optimization process for improving the visual quality of AIGODIs. We establish a comprehensive database named OHF2024 and conduct an extensive subjective experiment to collect detailed human feedback, including both human visual experience scores and distortion-aware saliency maps for AIGODIs.
We further propose two models with shared encoders based on BLIP-2 model and task-specific decoders to evaluate human visual experience and predict distortion-aware saliency based on the constructed database, which are named as BLIP2OIQA and BLIP2OISal, respectively.
Extensive experiments show that our BLIP2OIQA and BLIP2OISal models attain state-of-the-art performance on human visual experience evaluation and distortion-aware saliency prediction tasks for AIGODIs.
Finally, we present an effective optimization process, which utilizes the human visual experience scores and distortion-aware saliency predicted based on our proposed models to help enhance the visual quality of AIGODIs.

%references
%\begin{thebibliography}{1}
\vspace{-3pt}
\bibliographystyle{IEEEtran}
\small\bibliography{refs}
%\end{thebibliography}
\vspace{-33pt}
\begin{IEEEbiography}
[{\includegraphics[width=1in,height=1.25in,clip,keepaspectratio]{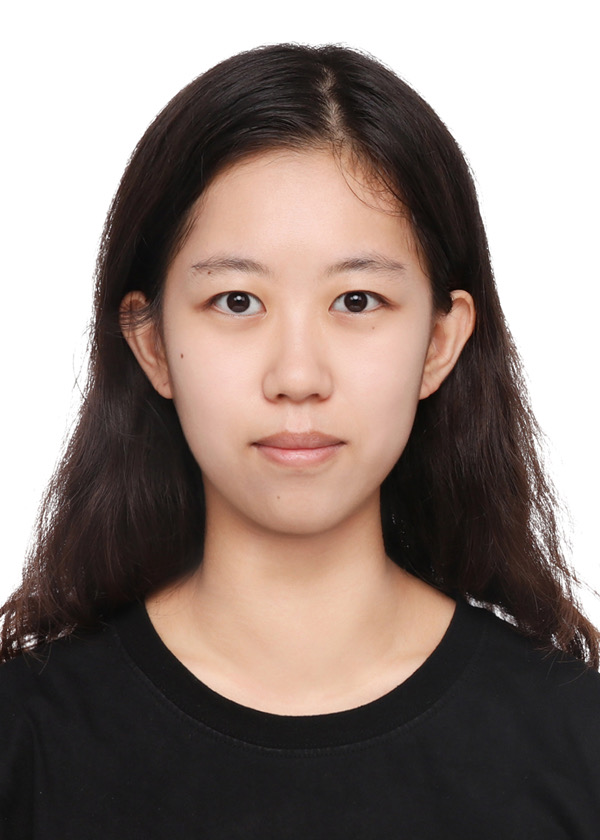}}]{Liu Yang}
received the B.E. degree from Shanghai Jiao Tong University, Shanghai, China, in 2024.
She is currently working toward the Master's degree
with the Department of Electronic Engineering,
Shanghai Jiao Tong University. Her research interests include perceptual quality assessment, quality
of experience and multimodal signal processing.
\end{IEEEbiography}
\vspace{-33pt}
\begin{IEEEbiography}
[{\includegraphics[width=1in,height=1.25in,clip,keepaspectratio]{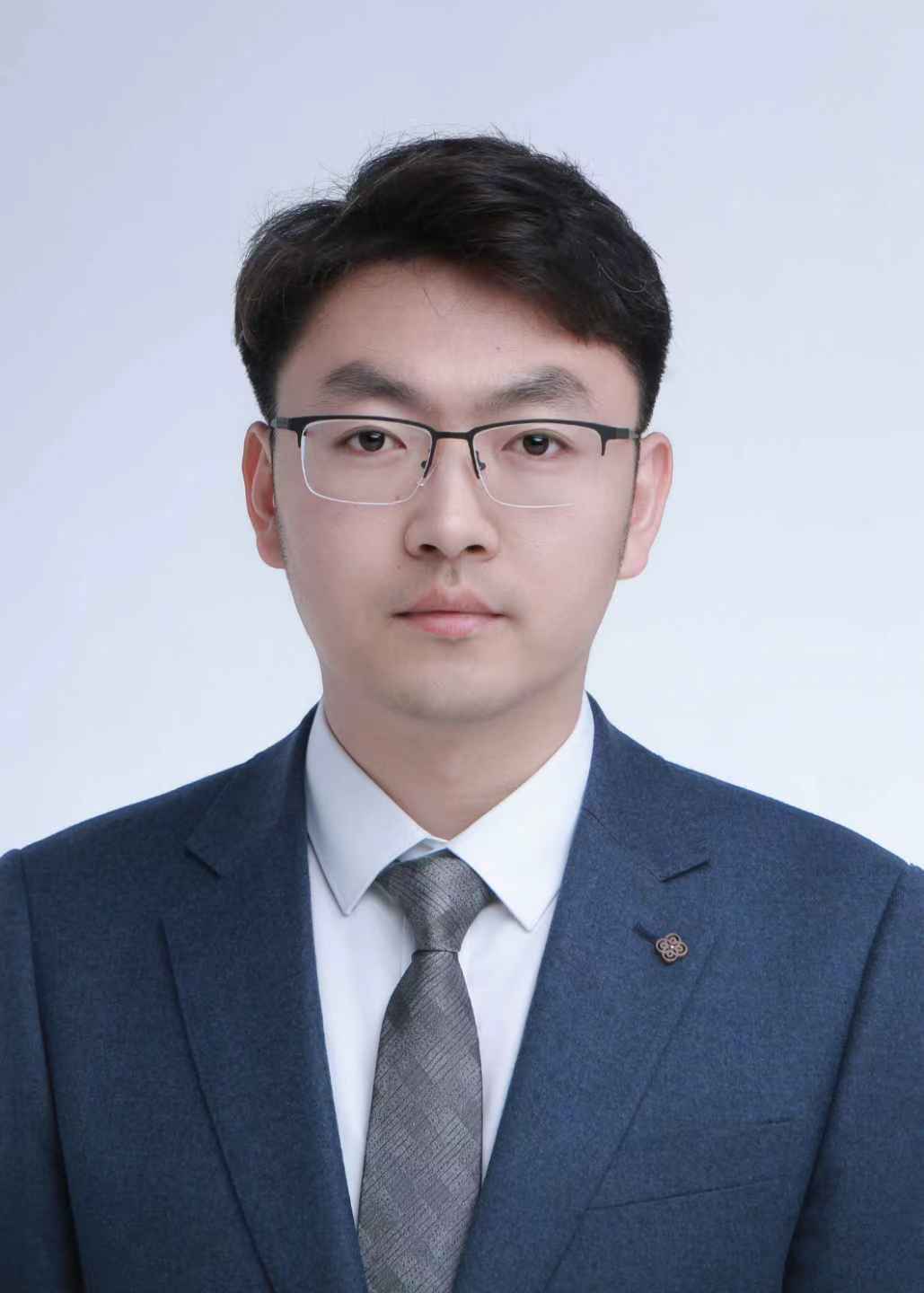}}]{Huiyu Duan}
received the B.E. degree from the University of Electronic Science and Technology of China, Chengdu, China, in 2017, and the Ph.D. degree from Shanghai Jiao Tong University, Shanghai, China, in 2024. He is currently a Postdoctoral Fellow at Shanghai Jiao Tong University. From Sept. 2019 to Sept. 2020, he was a visiting Ph.D. student at the Schepens Eye Research Institute, Harvard Medical School, Boston, USA. He received the Best Paper Award of IEEE International Symposium on Broadband Multimedia Systems and Broadcasting (BMSB) in 2022. His research interests include perceptual quality assessment, quality of experience, visual attention modeling, extended reality (XR), and multimedia signal processing.
\end{IEEEbiography}
\vspace{-33pt}
\begin{IEEEbiography}
[{\includegraphics[width=1in,height=1.25in,clip,keepaspectratio]{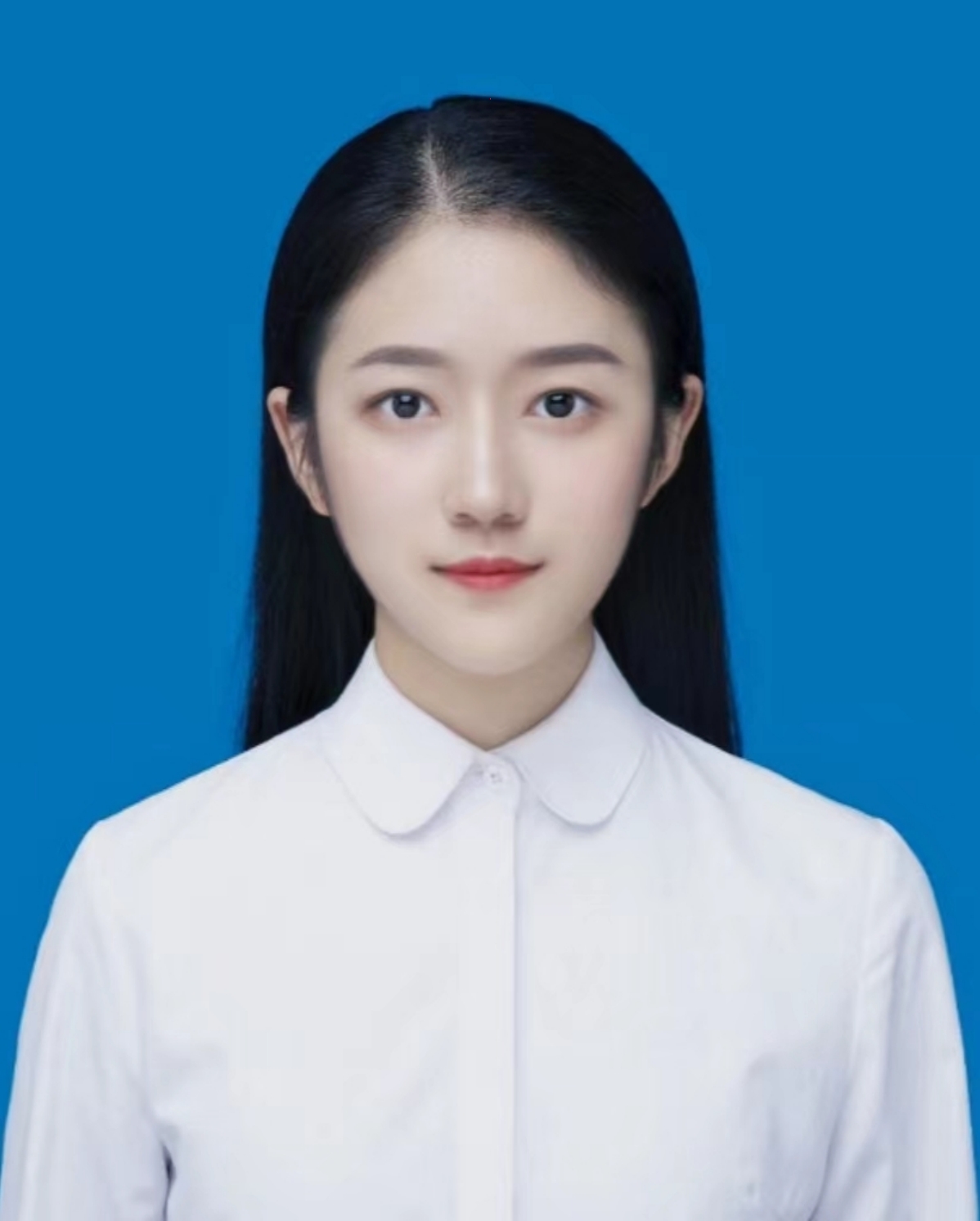}}]{Jiarui Wang}
received the B.E. degree from Shanghai Jiao Tong University, Shanghai, China, in 2024.
She is currently working toward the Ph.D. degree
with the Department of Electronic Engineering,
Shanghai Jiao Tong University. Her research interests include perceptual quality assessment, quality
of experience and multimodal signal processing.
\end{IEEEbiography}
\vspace{-33pt}
\begin{IEEEbiography}
[{\includegraphics[width=1in,height=1.25in,clip,keepaspectratio]{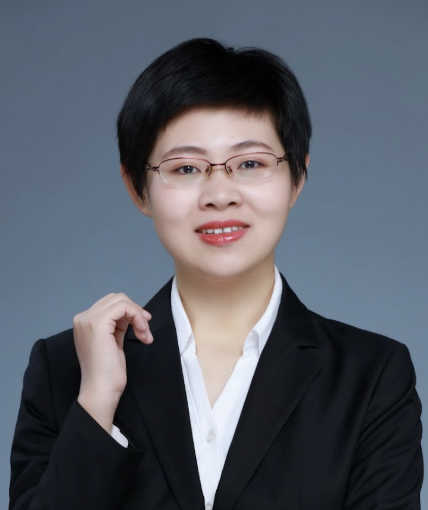}}]{Jing Liu}
(Member, IEEE) received the B.E. and Ph.D. degrees from Shanghai Jiao Tong University, Shanghai, China, in 2011 and 2017, respectively. From 2014 to 2015, she visited the Department of Computer Science and Engineering, State University of New York at Buffalo, USA. She is currently an Associate Professor with the Multimedia Institute, Tianjin University, Tianjin, China. She has authored more than 60 refereed articles and has filed ten patents. Her research interests include image/video processing and perceptual visual analysis. She was enrolled in “Beiyang Scholars-Young Teachers” Plan and received the Tianjin Science and Technology Progress Special Award in 2021.
\end{IEEEbiography}
\vspace{-33pt}
\begin{IEEEbiography}
[{\includegraphics[width=1in,height=1.25in,clip,keepaspectratio]{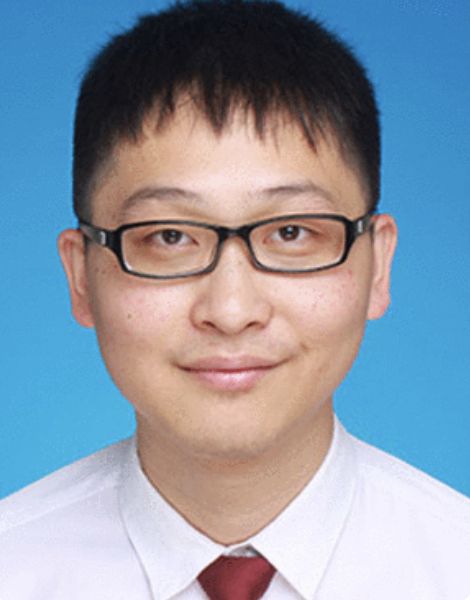}}]{Menghan Hu}
received the Ph.D. degree (Hons.) in biomedical engineering from the University of Shanghai for Science and Technology, Shanghai, China, in 2016.,From 2016 to 2018, he was a Postdoctoral Researcher with Shanghai Jiao Tong University, Shanghai. He is currently an Associate Professor with the Shanghai Key Laboratory of Multidimensional Information Processing, East China Normal University, Shanghai.
\end{IEEEbiography}
\vspace{-33pt}
\begin{IEEEbiography}
[{\includegraphics[width=1in,height=1.25in,clip,keepaspectratio]{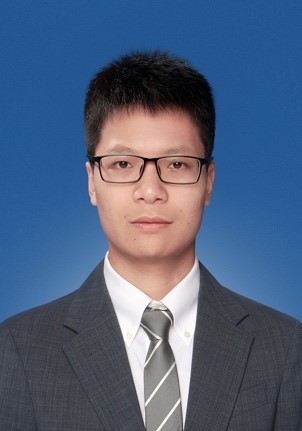}}]{Xiongkuo Min}
(Member, IEEE) received the B.E. degree from Wuhan University, Wuhan, China, in 2013, and the Ph.D. degree from Shanghai Jiao Tong University, Shanghai, China, in 2018, where he is currently a tenure-track Associate Professor with the Institute of Image Communication and Network Engineering. From Jan. 2016 to Jan. 2017, he was a visiting student at University of Waterloo. From Jun. 2018 to Sept. 2021, he was a Postdoc at Shanghai Jiao Tong University. From Jan. 2019 to Jan. 2021, he was a visiting Postdoc at The University of Texas at Austin. He received the Best Paper Runner-up Award of IEEE Transactions on Multimedia in 2021, the Best Student Paper Award of IEEE International Conference on Multimedia and Expo (ICME) in 2016, and the excellent Ph.D. thesis award from the Chinese Institute of Electronics (CIE) in 2020. His research interests include image/video/audio quality assessment, quality of experience, visual attention modeling, extended reality, and multimodal signal processing.
\end{IEEEbiography}
\vspace{-33pt}
\begin{IEEEbiography}
[{\includegraphics[width=1in,height=1.25in,clip,keepaspectratio]{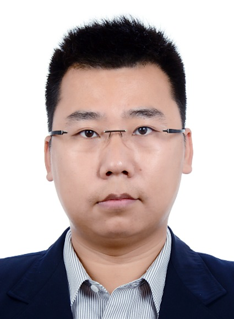}}]{Guangtao Zhai}
(F'24) received the B.E. and M.E. degrees from Shandong University, Shandong, China, in 2001 and 2004, respectively, and the Ph.D. degree from Shanghai Jiao Tong University, Shanghai, China, in 2009, where he is currently a Research Professor with the Institute of Image Communication and Information Processing. From 2008 to 2009, he was a Visiting Student with the Department of Electrical and Computer Engineering, McMaster University, Hamilton, ON, Canada, where he was a Post-Doctoral Fellow from 2010 to 2012. From 2012 to 2013, he was a Humboldt Research Fellow with the Institute of Multimedia Communication and Signal Processing, Friedrich Alexander University of Erlangen-Nuremberg, Germany. He received the Award of National Excellent Ph.D. Thesis from the Ministry of Education of China in 2012. His research interests include multimedia signal processing and perceptual signal processing.
\end{IEEEbiography}
\vspace{-33pt}
\begin{IEEEbiography}
[{\includegraphics[width=1in,height=1.25in,clip,keepaspectratio]{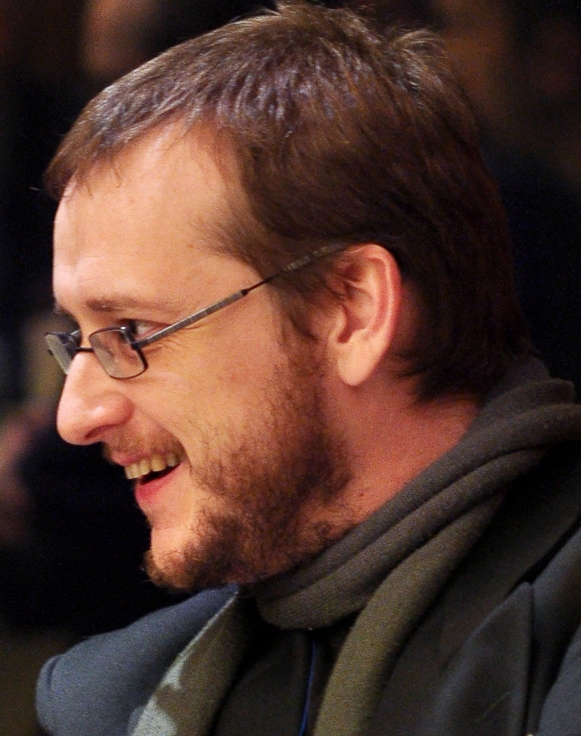}}]{Patrick Le Callet}
(F'19) received the M.Sc. and Ph.D. degrees in image processing from the Ecole Polytechnique de 1‘Universit\'{e} de Nantes. He was an Assistant Professor from 1997 to 1999 and a full time Lecturer from 1999 to 2003 with the Department of Electrical Engineering, Technical Institute of the University of Nantes. He led the Image and Video Communication Laboratory, CNRS IRCCyN, from 2006 to 2016, and was one of the five members of the Steering Board of CNRS, from 2013 to 2016. Since 2015, he has been the Scientific Director of the cluster Ouest Industries Cratives, a five-year program gathering over ten institutions (including three universities). Since 2017, he has been one of the seven members of the Steering Board of the CNRS LS2N Laboratory (450 researchers), as a Representative of Polytech Nantes. Since 2019, he has been the fellow of IEEE. He is mostly involved in research dealing with the application of human vision modeling in image and video processing.
\end{IEEEbiography}
\vfill

\end{document}